\documentclass{article}

\usepackage[final]{cpal_2025}

\usepackage{url}
\usepackage[dvipsnames]{xcolor}
\usepackage{hyperref}
\usepackage{amsmath}
\usepackage{bbm}
\usepackage{amssymb}
\usepackage{enumitem}
\usepackage{tikz}
\usetikzlibrary{arrows.meta, positioning, fit}
\usepackage{graphicx} 
\usepackage{subcaption}
\usepackage{booktabs}
\usepackage{float}

\usepackage{algorithm}
\usepackage{algpseudocode}

\title{Parameter-Efficient Distributional RL via Normalizing Flows and a Geometry-Aware Cramér Surrogate}

\author{%
  Simo Alami C.\textsuperscript{1},
  Rim Kaddah\textsuperscript{2},
  Marie-Paule Cani\textsuperscript{1},
  Jesse Read\textsuperscript{1}\\ 
  \textsuperscript{1}LIX, CNRS, Ecole Polytechnique, Institut Polytechnique de Paris,\\
  \textsuperscript{2}IRT SystemX\\
  \texttt{\{mohamed.alami-chehboune, marie-paule.cani, jesse.read\}@polytechnique.edu, rim.kaddah@irt-systemx.fr}
}

\begin{document}

\maketitle

\begin{abstract}
  Distributional Reinforcement Learning (DistRL) improves upon expectation-based methods by modeling full return distributions, but standard approaches often remain far from parsimonious. Categorical methods (e.g., C51) rely on fixed supports where parameter counts scale linearly with resolution, while quantile methods approximate distributions as discrete mixtures whose piecewise-constant densities can be wasteful when modeling complex multi-modal or heavy-tailed returns. We introduce NFDRL, a parsimonious architecture that models return distributions using continuous normalizing flows. Unlike categorical baselines, our flow-based model maintains a compact parameter footprint that does not grow with the effective resolution of the distribution, while providing a dynamic, adaptive support for returns. To train this continuous representation, we propose a Cramér-inspired, geometry-aware distance defined over probability masses obtained from the flow. We show that this distance is a true probability metric, that the associated distributional Bellman operator is a $\sqrt{\gamma}$-contraction, and that the resulting objective admits unbiased sample gradients—properties that are typically not simultaneously guaranteed in prior PDF-based DistRL methods. Empirically, NFDRL recovers rich, multi-modal return landscapes on toy MDPs and achieves performance competitive with categorical baselines on the Atari-5 benchmark, while offering substantially better parameter efficiency. 
\end{abstract}

\section{Introduction}

Traditional reinforcement learning (RL) algorithms aim to estimate the expected return from a given state or state-action pair \citep{sutton2018reinforcement}. However, this expectation provides only a partial view of the underlying return distribution, omitting critical information about uncertainty, risk, and variability. Distributional reinforcement learning (DistRL) addresses this limitation by modelling the full return distribution, providing a richer and more informative signal for decision-making.

\paragraph{Why distributional RL?} Modeling the full return distribution can improve learning and control beyond what is possible with expectation-only value functions. Prior work reports that distributional critics provide richer training signals and can yield stronger empirical performance, while also enabling risk-aware decision making through functionals of the return distribution (e.g., tail probabilities or CVaR) rather than a single mean value \cite{bellemare2017distributional,dabney2018implicit}. In addition, explicit access to distributional information supports robustness considerations and uncertainty-aware policies in stochastic environments, where variability and rare outcomes matter.

A critical challenge in DistRL is achieving an expressive representation of the return distribution without incurring excessive computational or parametric costs. Early approaches such as Categorical DQN (C51) \citep{bellemare2017distributional} approximate the return distribution using a histogram over a fixed discrete support. This formulation suffers from a fundamental lack of parsimony: the model's parameter count scales linearly with the support resolution. To achieve high precision, one must drastically increase the number of atoms, inflating the model size. Furthermore, the fixed support imposes a hard bound on returns, limiting applicability in domains with dynamic or highly variable rewards, such as robotic control \citep{horizon} or finance \citep{finance}.

Existing alternatives face theoretical or representational trade-offs. \citet{bellemare2017distributional} showed that the Distributional Bellman operator is a contraction in the Wasserstein metric. However, C51 is trained using KL divergence because empirical Wasserstein minimization suffers from biased sample gradients. This creates a disconnect between the contraction metric and the optimization objective. Moreover, using the KL divergence is also a major drawback in the unbounded setting, where predicted and target distributions may have negligible overlap, resulting in poor gradient signals. Subsequent quantile-based methods like QR-DQN \citep{dabney2017distributional} and IQN \citep{dabney2018implicit} or FQF \cite{FQF} address the support limitation by learning quantile locations and train with quantile regression losses (pinball, often with a Huber smoothing). By construction, quantile models handle unbounded returns and optimize a surrogate consistent with the 1-Wasserstein distance in one dimension. While highly effective, these methods approximate distributions as mixtures of Dirac deltas (step functions). This representation can be representationally inefficient for smooth, multi-modal, or heavy-tailed distributions, often requiring a large number of quantiles to approximate simple densities \citep{NC_quantile}.


A complementary approach is to model the return distribution through a continuous probability density. Continuity matters for two reasons. First, it yields an adaptive support and avoids committing to a fixed
return range, which is a limitation of categorical methods in domains with dynamic or heavy-tailed rewards. Second, a smooth density facilitates the computation of risk-sensitive criteria (e.g., CVaR or tail probabilities) by direct integration, without introducing an additional discretization level or increasing the number of parameters to refine resolution. This motivates continuous density models for DistRL. While recent work has explored normalizing flows for DistRL \citep{Theate_2023}, achieving a training objective that is both stable in RL and theoretically consistent (contraction and unbiased sample gradients) remains challenging.

\textbf{Contribution.} We propose NFDRL, a parameter-efficient DistRL method that models return distributions using continuous normalizing flows \citep{papamakarios2021normalizing}. To train this continuous representation, we introduce a geometry-aware surrogate of the Cramér distance defined over probability masses. Our approach offers three distinct advantages. First, it ensures \textbf{representational efficiency:} unlike categorical baselines where model size scales linearly with resolution, NFDRL maintains a compact, constant parameter footprint while offering infinite effective resolution and dynamic support. Second, it guarantees \textbf{theoretical soundness}; we prove that our mass-based distance is a true metric that induces a $\sqrt{\gamma}$-contraction in the distributional Bellman operator and admits unbiased sample gradients, resolving the inconsistencies of prior PDF-based approaches. Finally, the method demonstrates superior \textbf{expressivity}, capturing complex, multi-modal return landscapes that quantile baselines often fail to resolve.

We validate our approach on toy MDPs and the Atari-5 benchmark \citep{ATARI5}. We show that NFDRL achieves performance competitive with categorical baselines while utilizing significantly fewer parameters, offering a theoretically rigorous and parsimonious alternative for Distributional RL.

\paragraph{Scope and positioning.}
Our objective is not to propose a replacement for strong quantile-based agents such as IQN on Atari, but to study a parsimonious \emph{continuous} return representation with well-controlled metric properties. Quantile methods represent $\eta^\pi(x,a)$ as a finite mixture of Dirac masses; they are highly effective for control, but their discrete nature can make fine-grained characterization of distributional \emph{shape} (e.g., multimodality, tail behavior) dependent on the number of quantiles. In contrast, NFDRL models a smooth density with adaptive support and a constant parameter footprint, and is paired with a Cramér-inspired objective for which we prove (i) metric axioms, (ii) a $\sqrt{\gamma}$-contraction of the distributional Bellman operator, and (iii) unbiased sample gradients. We view NFDRL as a complementary tool with different trade-offs (parsimony, continuous PDFs, and theoretical consistency), particularly relevant when downstream risk-sensitive quantities depend on the detailed shape of the return distribution.

\section{Background}

We consider the standard RL setting, where the interaction between an agent and an environment is modelled as a Markov Decision Process (MDP) $\mathcal{M} = (\mathcal{X}, \mathcal{A}, R, P, \gamma)$ \citep{sutton2018reinforcement}. Here, $\mathcal{X}$ and $\mathcal{A}$ denote the state and action spaces, $R: \mathcal{X} \times \mathcal{A} \to \mathbb{R}$ is the reward function, $P(\cdot|x, a)$ is the transition kernel, and $\gamma \in (0, 1)$ is a discount factor. A policy $\pi(\cdot|x)$ maps a state $x \in \mathcal{X}$ to a distribution over actions. 

Under a fixed policy $\pi$, the return $G^\pi(x)$ is defined as the random variable representing the discounted sum of rewards collected along a trajectory starting from state $x$:
\begin{equation}G^\pi(x) := \sum_{t=0}^\infty \gamma^t R(x_t, a_t), \quad x_0 = x, a_t \sim \pi(\cdot|x_t), x_{t+1} \sim P(\cdot|x_t, a_t).\end{equation}

The goal is to estimate the expected return i.e. the action-value function $Q$: 

\begin{equation*} Q^\pi(x, a) := \mathbb{E} [G^\pi(x, a)] = \mathbb{E} \left[ \sum_{t=0}^\infty \gamma^t R(x_t, a_t) ,\Big|, x_0 = x, a_0 = a \right].\end{equation*}

These quantities satisfy the Bellman equation:
\begin{equation}Q^\pi(x, a) = \mathbb{E}[R(x, a)] + \gamma \mathbb{E}_{X' \sim P(\cdot|x, a), A' \sim \pi(\cdot|X')} [Q^\pi(X', A')].\end{equation}

\subsection*{Distributional Reinforcement Learning}
Rather than approximating only the expectation of returns, DistRL aims to model the entire distribution of $G^\pi(x)$, capturing richer information such as variance, multimodality, or higher moments. This has been shown to improve both learning dynamics and empirical performance \citep{bellemare2017distributional}.



We adopt the formulation proposed by \citet{bdr2023}, which distinguishes between the random return $G^\pi(x)$ and its law $\eta^\pi(x)$, defined as:
\begin{equation}\eta^\pi(x)(S) := \mathbb{P}(G^\pi(x) \in S), \quad \forall S \subseteq \mathbb{R}.\end{equation}

To express the Bellman recursion at the level of distributions, they introduce the concept of a pushforward function. Given a function $f: \mathbb{R} \to \mathbb{R}$ and a probability distribution $\eta$, intuitively we have,  if $Z \sim \eta$, then $f(Z) \sim f_\#\eta$.

Let the bootstrap function defined as $b_{r,\gamma}: \mathbb{R} \to \mathbb{R}$ by $b_{r,\gamma}(z) := r + \gamma z$, for fixed $r \in \mathbb{R}$ and $\gamma \in (0, 1)$. Then the distributional Bellman equation over distributions can be written as:
\begin{equation}\eta^\pi(x) = \mathbb{E}_{a \sim \pi(\cdot|x), X' \sim P(\cdot|x,a)} [(b_{R(x,a), \gamma})_\# \eta^\pi(X')].\end{equation}

This defines the distributional Bellman operator $\mathcal{T}^\pi: \mathcal{P}(\mathbb{R})^\mathcal{X} \to \mathcal{P}(\mathbb{R})^\mathcal{X}$, where:
\begin{equation}\label{eq:bootstrap_Bellman}
(T^\pi \eta)(x) := \mathbb{E}_{a \sim \pi(\cdot|x), X' \sim P(\cdot|x,a)} [(b_{R(x,a), \gamma})_\# \eta(X')].
\end{equation}

This approach provides a compact, principled notation for distributional TD methods we use.

\subsection*{Normalizing Flows}
NF are a class of generative models that transform a simple base distribution into a more complex one via a sequence of smooth, invertible mappings \citep{papamakarios2021normalizing}. Starting from a base sample $z_0 \sim p(z_0)$, a flow applies a sequence of transformations $f_k \circ \dots \circ f_1$ to obtain $z_K$, whose density is computed using the change of variables formula:
\begin{equation}\label{eq:NF_general}
\log p_\theta(z_K) = \log p(z_0) - \sum_{k=1}^K \log \left| \det \left( \frac{\partial f_k}{\partial z_{k-1}} \right) \right|.
\end{equation}
This allows flows to model flexible densities while retaining exact likelihoods and differentiability. 

\section{Normalizing Flows for Distributional RL}

We present our NF-based model for return distribution estimation. While normalizing flows can map in either direction—base to return distribution or vice versa—RL constraints guide this choice. We first discuss both options and justify the appropriate direction, then describe the model architecture and how the distributional Bellman operator is implemented using pushforward distributions.

\subsection{Modeling the Forward Flow: Sampling Returns and Evaluating Densities}

Let $\mathcal U$ be a base distribution with full support over $\mathbb{R}$, and let $z \sim \mathcal U$. Let $f$ be a diffeomorphism from $\mathbb{R}$ to itself (i.e., a bijective, differentiable mapping with a differentiable inverse). 

The flow $f$ can be interpreted in two equivalent ways, (i) either as mapping return outcomes $y\sim \eta^\pi(x,a)$ into the base space: $z=f(y)$, or (ii) as mapping base samples $z\sim \mathcal U$ to return outcomes via: $y=f(z)$, where $y\sim\eta^\pi(x,a)$. 

Mapping from return outcomes to base samples (i.e., modeling the inverse flow) is attractive because it allows direct likelihood evaluation of observed returns, which is beneficial for maximum likelihood training. However, in RL, we do not observe true return samples $y\sim\eta^\pi(x,a)$ directly. We must construct them through bootstrapping. This makes the inverse mapping impractical in our context. Instead, we opt to model the forward flow, transforming base noise samples $z\sim\mathcal U$ into return outcomes $y=f_\theta(z)$, which allows us to generate samples from $\eta^\pi(x,a)$ and define pushforward distributions suitable for implementing the distributional Bellman operator.

\begin{figure}[H]
    \centering
    \resizebox{0.89\textwidth}{!}{%
    \scalebox{0.7}{ 
\begin{tikzpicture}[->, >=Stealth, node distance=2cm]
\tikzstyle{mynode}=[thick,draw=blue,fill=blue!20,circle,minimum size=22]
\tikzstyle{opnode}=[rectangle, rounded corners, draw=Apricot, fill=Apricot!50, minimum size=0.8cm]
\tikzstyle{state}=[circle, draw=MidnightBlue, fill=MidnightBlue!30, minimum size=1.2cm]
\tikzstyle{inputnode}=[rectangle, draw=Emerald, fill=Emerald!20, minimum size=0.8cm]

\node[inputnode] (X) {$X$};
\node[opnode] (h) [right of=X] {$h_\theta$};
\node[state] (w) [above right of=h] {$w_i^j$};
\node[state] (m) [right of=h] {$\mu_i^j$};
\node[state] (s) [below right of=h] {$\sigma_i^j$};
\node[opnode] (F) [right of=m] {$F^{(a_j)}(z_k)$};
\node[opnode] (F') [below of=F] {$\frac{\partial F^{(a_j)}(z_k)}{\partial z_k}$};
\node[inputnode] (z) [above of=F] {$z_k\sim \mathcal{U}$};
\node[state] (y) [right=2cm of F] {$y_k^j$};
\node[state] (p) [right=2cm of F'] {$\eta^{\pi}(y_k^j)$};
\node (dummy) [right=3cm of y] {};
\node[opnode] (g) [above right=2cm of F] {$b_{R,\gamma}^{X'}(y_k^j)$};
\node[inputnode] (X') [above left of=g] {$X', R$};
\node[opnode] (g') [above of= g] {$\frac{\partial g(y_k^j)}{\partial y_k^j}$};
\node[state] (y') [right of=g] {$\hat{y}_k^j$};
\node[state] (p') [right of=g'] {$T^\pi\eta(\hat{y}_k^j)$};
\node (dummy1) [above of=X'] {};

\draw (X) -- node[right] {} (h);
\draw (h) -- node[above right] {} (w);
\draw (h) -- node[right] {} (m);
\draw (h) -- node[below right] {} (s);
\draw (w) -- node[below right] {} (F);
\draw (m) -- node[right] {} (F);
\draw (s) -- node[above right] {} (F);
\draw (z) -- node[below] {} (F);
\draw (F) -- node[above right] {} (y);
\draw (F) -- node[below] {} (F');
\draw (F') -- node[right] {} (p);
\node[draw, dashed, fit=(y)(p)(dummy), inner sep=5pt, label={[anchor=north east]north east:Predicted distribution}] (group) {};
\draw (F) -- node[above right] {} (g);
\draw (F') -- node[above right] {} (g);
\draw (X') -- node[below right] {} (g);
\draw (g) -- node[above] {} (g');
\draw (g) -- node[right] {} (y');
\draw (g') -- node[right] {} (p');
\node[draw, dashed, fit=(X')(g)(g')(y')(p')(dummy1), inner sep=5pt, label={[anchor=north east]north east:Target distribution}] (group) {};

\end{tikzpicture}
}}
    \caption{Architecture of the conditional flow model. A neural network $h_\theta$ maps each state $x$ to parameters $\{(w_i^j, \mu_i^j, \sigma_i^j)\}_{i=1}^n$ for each action $a_j\in\mathcal{A}$, defining a mixture CDF $F^{(a_j)}$. This CDF acts as a flow that transforms base noise samples $z_k\sim\mathcal{U}$ into return samples $y_k^j = F^{(a_j)}(z_k)$. The change of variable formula then makes use of the flows derivative to approximate the return distribution $\eta^\pi(y_k^j)$. To estimate the target distribution, the bootstrap function is implemented as a flow and takes the reward and next state as input. It finally outputs target distribution $T^\pi\eta$.}
    \label{fig:architecture}
\end{figure}

While choosing to represent return samples as $y=f_\theta(z)$, with $z\sim\mathcal U$, makes sampling straightforward, it is also possible to compute the probability density of a given return value using the change of variable formula. Let $f_\theta:\mathbb R\xrightarrow{}\mathbb R$ be a flow parameterized by $\theta$,  and $p_{\mathcal{U}}$ the density of the base distribution $\mathcal U$. Then for $y=f_\theta(z)$, we have:
\begin{equation}\label{eq:change_variable}
   \log \eta^\pi(x,a)(y) = \log\left(p_{\mathcal{U}}(z)\left|\frac{\partial y}{\partial z}\right|^{-1}\right)
   = \log\left(p_{\mathcal{U}}(z)\left|\frac{\partial f_\theta(z)}{\partial z}\right|^{-1}\right)
   = \log p_{\mathcal{U}}(z) - \log\left|\frac{\partial f_\theta(z)}{\partial z}\right|
\end{equation}

This formulation yields a closed-form expression for the learned density. Computing the log-density of sampled returns requires only the base log-density and the derivative of the flow function. Choosing a flow function with easily computable derivatives is crucial.

\subsection{A CDF-Based Flow Architecture for Conditional Return Modeling}\label{sec:CDF}

We propose an architecture in which a conditional flow function, constructed from a CDF, maps base samples to return values. Since return distributions $\eta^\pi(x,a)$ are one-dimensional, we design a 1D flow $F_\theta(x, a): \mathcal{Z} \rightarrow \mathbb{R}, \text{ where } z$ is drawn from a fixed base distribution, and y = $F_{\theta(x, a)}(z)$ is a return sample conditioned on the state-action pair $(x,a)$.

To model this conditional transformation, we use a neural network $h_\theta$ that takes as input the state $x$ and outputs, for each discrete action $a_j\in\mathcal{A}$, the parameters of a Gaussian mixture distribution: $\{(w_i^{(j)}, \mu_i^{(j)}, \sigma_i^{(j)})\}_{i=1}^n$. These parameters define a mixture CDF, and with $z\sim\mathcal{U}$, we get:
\begin{equation*}
    y^{a_j} = F^{(a_j)}(z) = \sum_{i=1}^n w_i^{(j)} \Phi\left(\frac{z - \mu_i^{(j)}}{\sigma_i^{(j)}}\right)
\end{equation*}


\paragraph{Why CDF-based flows?}
Since $\eta^\pi(x,a)$ is one-dimensional, we adopt a CDF-based parameterization to balance expressivity and numerical stability. Concretely, we model the conditional transformation through a Gaussian-mixture CDF, which is strictly monotone by construction and whose derivative is a Gaussian-mixture PDF. This yields closed-form density evaluation via the change-of-variables formula while retaining a flexible, non-linear family in 1D.

Although this CDF does not admit a closed-form inverse in general, strict monotonicity enables robust inversion by 1D binary search when evaluating likelihoods. In our setting, the forward map $z \mapsto y$ is cheap and is the only operation required for sampling-based Bellman updates; the inverse is used only for density evaluation and can be computed reliably with logarithmic-time search. Finally, we emphasize that this design choice is motivated by tractability in the 1D return setting rather than by universal superiority over other flow families (e.g., affine/squared flows or spline flows). Different parameterizations may be preferable under different computational or modeling constraints; we therefore present Gaussian-mixture CDF flows as a simple and expressive choice well suited to our use case.

In contrast, our CDF-based approach naturally supports smooth, continuous outputs and avoids discretization altogether. It is conceptually closest to the method in \citep{ho}, which uses a logistic mixture CDF composed with an inverse sigmoid. Like ours, it supports expressive, invertible transformations without sacrificing differentiability.

\subsection{Rescaling the Flow Output: Addressing the Bounded Support of CDFs}

Because the chosen flow function is based on a CDF, its output is restricted to $[0, 1]$. However, one of our goals is to allow the predicted return distribution to be continuous and adaptable, reflecting the nature of real-world return values in RL. To overcome this limitation, we introduce an additional transformation step that rescales the output of the CDF-based flow. 

$h_\theta$ is extended to output an additional parameter $G^{\max}_{(x,a)}$, which defines the upper bound of the return range for a given state-action pair. The model can hence flexibly adapt the support of the predicted distribution without manual specification. In this case, we apply an affine transformation to map the CDF output $y \in [0,1]$ to the desired return range: $f(y) = 2\cdot y \cdot G^{max} - G^{max}$. The function $f$ is considered as a flow function that comes after the first CDF flow $F$.


We are now composing two flows ($F$ and $f$) and as per \eqref{eq:NF_general}, this transformation introduces an additional Jacobian term in the log-likelihood of \eqref{eq:change_variable}:
\begin{equation}
\log \eta^\pi(x,a)(y) = \log p_{\mathcal{U}}(z) - \log \left| \frac{dF_\theta(z)}{dz} \right| - \log |2\cdot G^{max}|.
\end{equation}

\subsection{Constructing the RL Target Distribution}\label{sec:target}

Before proceeding, we clarify terminology: in normalizing flows, the target distribution is the flow's output. In our RL setting, this is the predicted return distribution for a state-action pair. During training, it is compared to a separate RL target distribution, which serves as the learning signal. Unless otherwise noted, target distribution refers to this RL target.

We begin by reviewing how target distributions are commonly constructed in DistRL. We then show why directly applying existing methods to our flow-based model does not yield a valid distribution. Finally, we propose a principled solution by introducing a \textit{target flow} that enables a coherent learning objective within our framework.

As shown in ~\eqref{eq:bootstrap_Bellman}, the Bellman operator $T^\pi$ applies the bootstrap function $b_{r, \gamma}(y) = r + \gamma y$ to samples $y = F_\theta(z)$. However, this scaling distorts probability mass, breaking the change-of-variables formula. \textbf{Our key contribution is to treat $b_{r,\gamma}$ as a flow layer}. Being affine and invertible, it integrates naturally into the model, preserving normalization. The full composed flow then yields a valid target distribution under the change-of-variable formula.

Let $F_{x',a'}$ denote the flow that predicts the return distribution for the next state-action pair,  and let $\tilde{y}=g(y) = r + \gamma y$ be the bootstrap flow. We now have a composition of 3 flows ($F$, $f$ and $g$) and applying \eqref{eq:NF_general}, The target log-density becomes:

\vspace{-1.5\baselineskip}
\begin{align}
\log T^\pi\eta(x,a)(\tilde{y}) 
&= \log p_z(z) 
   - \log \left| \frac{\partial F_{x',a'}(z)}{\partial z} \right|- \log |2\cdot G^{max}| - \log \left| \frac{\partial g(F_{x',a'}(z))}{\partial F_{x',a'}(z)} \right| \\
&= \log p_z(z) 
   - \log \left| \frac{\partial F_{x',a'}(z)}{\partial z} \right|
   - \log |2\cdot G^{max}| - \log(\gamma).
\label{eq:target_bootstrap}
\end{align}
\vspace{-1.5\baselineskip}

This approach guarantees that the target distribution is properly normalized, thanks to the compositionality of flows and the use of the change-of-variable formula. Additionally, the extra terms $\log(\gamma)$ and $\log |2\cdot G^{max}|$ are constant with respect to $z$ and do not incur computational overhead.

Applying the target flow alone does not yield a complete RL target distribution suitable for training, as the predicted distribution $\eta^\pi(x,a)$ and the target distribution $T^\pi \eta(x,a)$ are defined over different supports. As in the C51 algorithm, which requires projecting the Bellman update onto a fixed categorical support, we must ensure alignment between the two distributions to enable valid comparisons. Hence, we introduce an alignment procedure based on kernel density estimation (KDE). More specifically, we use a KDE to evaluate $\eta^\pi$ on the support of $T^\pi\eta(\tilde{y})$,  this is denoted as $\hat\eta^\pi$. The same is done for $T^\pi\eta$ evaluated on the support of $\eta^\pi$ (denoted $\hat T^\pi\eta$)\footnote{This is an abuse of notation as $\eta^\pi$ and $\hat\eta^\pi$ are the same distribution but evaluated on a different set of points. However as both are obtained using different processes; one is the direct output of the model and the other is obtained using a KDE, we use this notation to mark the difference.}. This alignment allows to estimate the target probability masses $v_i$ required for the loss defined in Section 3.5.

We then have:
\begin{align}\label{eq:loss_full}
\mathcal L(\eta^\pi(x,a),& T^\pi \eta(x,a)) = \sum_{i=1}^N D\!\left(\eta^\pi(x, a)(y_i) \;\middle\|\; 
   \hat{T}^\pi \eta(x, a)(y_i)\right) 
+ \sum_{j=1}^M D\!\left(\hat{\eta}^\pi(x, a)(\tilde{y}_j) \;\middle\|\; 
   T^\pi \eta(x, a)(\tilde{y}_j)\right).
\end{align}

Most importantly, as detailed in the appendix section \ref{KDE_appendix}, the KDE is an essential tool allowing our proposed loss (eq. \eqref{eq:loss_full}) and metric (eq.\eqref{eq:loss}) to behave like a transport distance.  

\paragraph{Final state Gaussian reward approximation} In standard TD methods, the value target at the final time step is simply the immediate reward $r$. In the distributional RL setting, however, we require a full target distribution. Since $r$ is a scalar, it can be viewed as a degenerate distribution — a Dirac delta centered at $r$. To enable learning with continuous distributions, we approximate this Dirac using a Gaussian $\mathcal{N}(r, \sigma)$. The alignment process is illustrated in the appendix figure 2 and detailed in appendix algorithm 1.

\subsection{Using the Cramér Distance as a Loss Function}

Unlike C51, which uses KL divergence, our method operates on adaptable supports where KL is ill-suited. KL fails when predicted and target distributions do not overlap—common early in training—and lacks translation sensitivity, making it ineffective for disjoint supports. While the Bellman operator is a contraction under the maximal form of the Wasserstein metric \citep{bellemare2017distributional}, this metric is hard to optimize with stochastic gradients. Quantile regression \citep{dabney2017distributional} addresses this for the 1-Wasserstein case but requires learning quantile values, incompatible with our full-PDF approach. Instead, we use the Cramér distance, which is translation-sensitive, invariant to mass-preserving transforms while providing unbiased sample gradients—making it ideal for our setting \citep{bellemare2017cramer}. Definitions and key properties are recalled in the appendix.

Recall that for two distributions with CDFs $F(x)$ and $G(x)$, the squared Cramér distance is:
\begin{equation*}
    l_2^2(P,Q) = \int_{-\infty}^{\infty} (F(x)-G(x))^2 \, dx
\end{equation*}

Computing the Cramér distance necessitates the CDFs of the predicted and target distributions which are not available. While using the available PDFs of the predicted and target distributions, it is possible to compute the CDFs then the exact Cramér distance; however the involved sorting operation on returns generates overhead and becomes tricky or ill-defined in higher dimensions, because multivariate CDFs are not straightforward, and ordering does not generalize well.

We propose a surrogate for the Cramér distance that is computed over the support of the distributions. Crucially, to preserve the contraction property of the Bellman operator, our loss is defined over probability masses rather than raw densities. We propose the following geometry-aware loss:
\begin{equation}\label{eq:loss}
\begin{aligned}
    D(\eta^\pi(x,a),\hat{T}^\pi \eta(x,a))
    &= \Biggl(\frac{1}{N^2}\sum_{i,j}
    \bigl(w_i - v_i\bigr)^2 \cdot |y_i - y_j| \Biggr)^{1/2}
\end{aligned}
\end{equation}
where $w_i$ and $v_i$ represent the discrete probability masses of the predicted and target distributions at support point $y_i$, respectively. This distinction is critical for theoretical convergence: when the return scale changes (e.g., by the discount factor $\gamma$), the support density scales inversely ($\propto 1/\gamma$) while the interval width scales proportionally ($\propto \gamma$). Consequently, the probability mass $w_i$ remains invariant.

These masses can be obtained via a Riemann approximation linking the continuous PDF to the discrete support: $w_i \approx \eta^\pi(x,a)(y_i) \cdot \Delta y$ and $v_i \approx \hat T^\pi\eta^\pi(x,a)(y_i) \cdot \Delta y$. In the specific case where the support is generated via Monte Carlo sampling (as in our base distribution sampling), each point is implicitly assigned a constant mass of $1/N$, which is inherently scale-invariant.

This formulation, using the linear transport cost $|y_i - y_j|$, retains the scale sensitivity of the original Cramér distance ($l_2(cX,cY) = \sqrt{c} \, l_2(X,Y)$) and allows for stochastic optimization with unbiased gradients. In Appendix sections D and E, we prove that this surrogate acts as a valid distance function and preserves contraction properties under the distributional Bellman operator.

\section{Results}

In this section, we empirically validate the proposed DistRL method (NFDRL). We illustrate its ability to model return distributions with tunable variance, showcasing the flexibility offered by our parameterization and the effect of surrogate optimization. We then investigate more complex settings, demonstrating the ability to learn multi-modal distributions. A particular focus is placed on a simple MDP taken from \citep{expectile_DRL} with a bimodal final state distribution (see Appendix figure 4), where quantile-based methods struggle to learn smooth distributions. In contrast, our method successfully handles this challenge. 

Building on these controlled experiments, we evaluate our method on a discrete stochastic environment, FrozenLake, to better visualize the multimodal distributions output by our model. We then scale to larger benchmarks by reporting results on a selection of Atari 2600 games, establishing the practicality of our approach in deep reinforcement learning settings.

Finally, we compare C51 and NFDRL model size and provide an empirical analysis of the sample size from base distribution to ensure effective training. 

\subsection{Modeling Expressiveness using simple MDPS}

\paragraph{Toy MDPs} We evaluate expressiveness on two controlled MDPs (setup detailed in Appendix C). In the deterministic MDP1, NFDRL demonstrates tunable granularity (Figure~\ref{fig:final}). While the surrogate loss yields slightly wider distributions than the exact Cramér loss—a consequence of minimizing a conservative upper bound over discrete masses—it crucially retains robust gradient signals via the geometry-aware term $|y_i-y_j|$. This avoids the vanishing gradients of KL-based methods on disjoint supports while mitigating the artificial broadening of C51 or the Dirac-mixture limitations of quantile baselines (see Appendix sections F and G). Additionally, the stochastic MDP2 confirms that NFDRL accurately captures multi-modal dynamics, recovering clear bimodal distributions (Figure~\ref{fig:bimodal}) where unimodal or quantile approximations often blur the density.

\begin{figure*}[t]
    \centering
    \includegraphics[width=\textwidth]{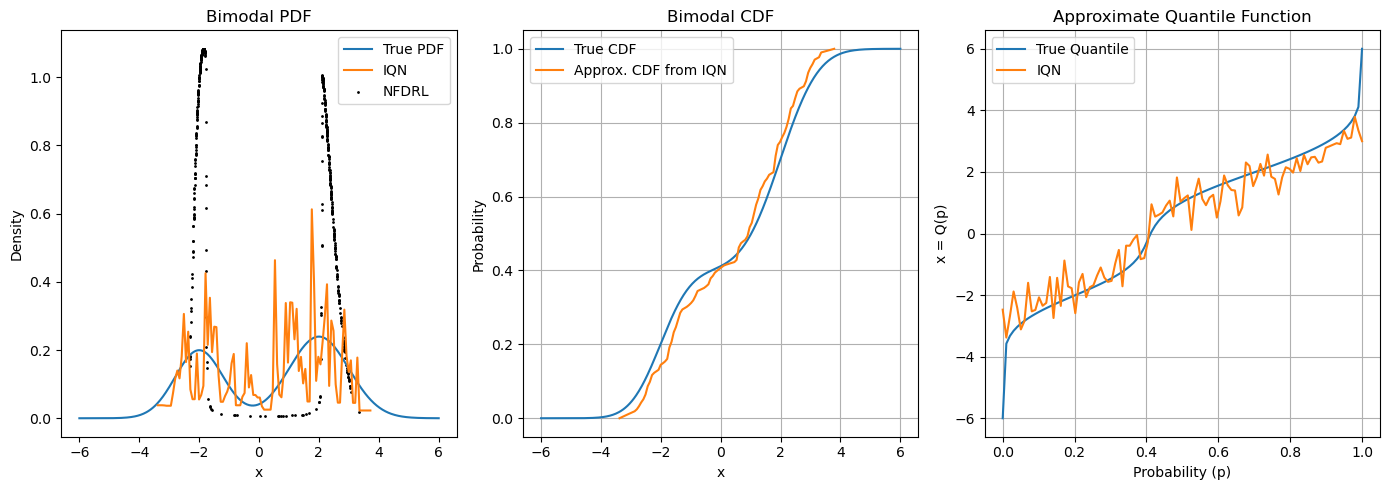}
    \caption{Return distributions learnt for the final state of $MDP_3$ using IQN and NFDRL. 
    \textbf{Right}: true quantile function (blue) and quantile function learnt using IQN, reproducing the results of \citep{expectile_DRL}. 
    \textbf{Middle}: True CDF and CDF approximated for the quantile function obtained from IQN. 
    \textbf{Left}: True PDF, and approximations obtained using IQN and NFDRL. 
    While IQN produces a noisy distribution blurring the models, our method outputs a smoother distribution that makes the modes apparent.}
    \label{fig:IQN_bimodal}
\end{figure*}

\paragraph{MDP3} We reproduce the MDP described in \citep{expectile_DRL} consisting of 4 successive states with only one possible action (Appendix Figure 4). The reward is nil for all states except the last where $R \sim \left(\frac{1}{2} \mathcal{N}(-2, 1) + \frac{1}{2} \mathcal{N}(+2, 1)\right)$. Figure \ref{fig:IQN_bimodal} displays the distributions learnt for that final state return using IQN and NFDRL. As noted in \citep{expectile_DRL}, quantile regression approximates the inverse CDF with high variance, especially at extremes, leading to noisy and blurred PDF estimates that obscure bimodality. In contrast, our method produces smooth, clearly bi-modal PDFs, albeit with sharper peaks.

\paragraph{Frozen Lake}
To further assess the expressiveness of our model, we evaluate it on the Frozen Lake environment ~\citep{gym}, a grid-world domain with inherent randomness. In figure \ref{fig:frozen_lake}, we display the learned return distributions for each state-action pair. The environment’s stochasticity naturally induces multimodal and skewed return distributions, which our model captures accurately.

\subsection{Benchmarking}

\paragraph{Environments} We opted to conduct our experiments with the Atari Learning Environment (ALE) \citep{ALE} and specifically the Atari-5 sub-benchmark \citep{ATARI5}. We also report human-normalized scores. Implementation details can be found in the appendix (section H).  

\paragraph{Results} We evaluate our proposed method, NFDRL, in both its exact (NFDRL-E) and surrogate (NFDRL-S) variants across five Atari games, comparing against established baselines including DQN, C51, and IQN. Performance is measured using human-normalized scores (table \ref{tab:human_normalized_scores}), computed based on the raw game scores of human players and random agents (Appendix table 1). Both NFDRL variants significantly outperform traditional baselines (DQN and C51) on all games, achieving mean scores of 394 (NFDRL-E) and 407 (NFDRL-S) versus 189 (DQN) and 320 (C51). Notably, NFDRL-S, which uses a surrogate Cramér loss, slightly outperforms the exact version on average, highlighting its favorable tradeoff between computational efficiency and performance.

Although NFDRL-S uses a surrogate Cramér loss, its slightly better performance compared to the exact version is consistent with trends in deep learning. The exact Cramér loss introduces sampling noise, discretization errors, and sensitivity to distribution tails, which can destabilize training. In contrast, the surrogate loss may better handle these challenges, implicitly reweighting or smoothing the distribution, making it more amenable to optimization. 

While IQN remains the top-performing method overall with a mean score of 525, our method performs competitively, particularly in games such as Double Dunk and QBert*, where NFDRL-S achieves or exceeds IQN performance. In sum, these results validate that the Cramér-based approach is a powerful alternative to quantile-based distributional RL, capable of modeling complex return distributions while maintaining strong empirical performance.

We emphasize that NFDRL is not intended to outperform IQN on Atari, but to provide a parsimonious continuous alternative with different trade-offs: constant parameter footprint and adaptive support, together with a metric-matched objective enjoying contraction and unbiased-gradient guarantees. This perspective is especially relevant in settings where decisions depend on the \emph{shape} of $\eta^\pi(x,a)$ (e.g., multimodality or tail risk), rather than only on expected return.

\begin{table}[h!]
\centering
\begin{tabular}{lccccc}
\toprule
\textbf{Games} & \textbf{DQN} & \textbf{C51} & \textbf{IQN} & \textbf{NFDRL-E} & \textbf{NFDRL-S} \\
\midrule
Battle Zone   & 79  & 76  & \textcolor{blue}{\textbf{115}} & 87  & \textbf{95}  \\
Double Dunk   & 545 & 959 & 1100 & 1200 & \textcolor{blue}{\textbf{1243}} \\
Name this Game & 103 & 178 & \textcolor{blue}{\textbf{354}} & 232 & \textbf{259} \\
Phoenix       & 119 & 258 & \textcolor{blue}{\textbf{862}} & 280 & \textbf{301} \\
Q*Bert        & 97  & 178 & 193 & 169 & \textcolor{blue}{\textbf{193}} \\
\midrule
\midrule
\textbf{Mean} & 189 & 330 & \textcolor{blue}{\textbf{525}} & 394 & \textbf{418} \\
\bottomrule
\end{tabular}%
\caption{Human-normalized performance on ATARI-5. The best performing agent is highlighted in blue. We also compare our approach directly with C51 as they are both approximating the PDF (best one written in bold font) while IQN is quantile based.}
\label{tab:human_normalized_scores}
\end{table}

\subsection{Parameter efficiency}

\begin{figure}[h!]
    \centering
    \begin{subfigure}[t]{0.48\columnwidth}
        \centering
        \includegraphics[width=\linewidth]{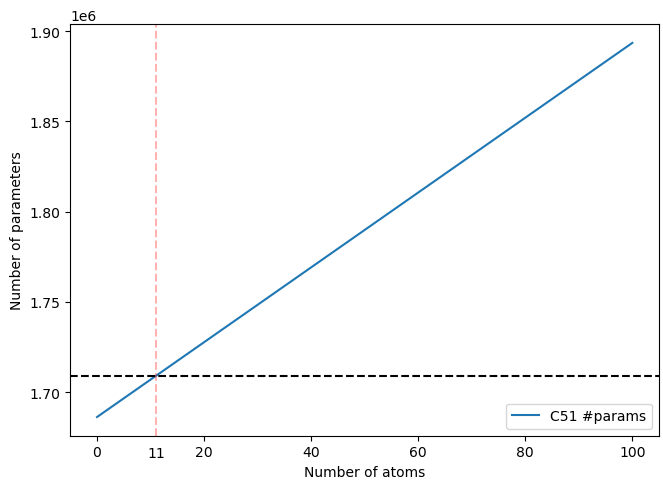}
        \caption{Parameter count comparison between C51 and NFDRL. 
        C51’s parameter count increases with the number of atoms due to its linear output layer. 
        NFDRL maintains a fixed number of parameters, matching C51 only when it uses 11 atoms, 
        thus demonstrating superior parameter efficiency.}
    \end{subfigure}
    \hfill
    \begin{subfigure}[t]{0.48\columnwidth}
        \centering
        \includegraphics[width=\linewidth]{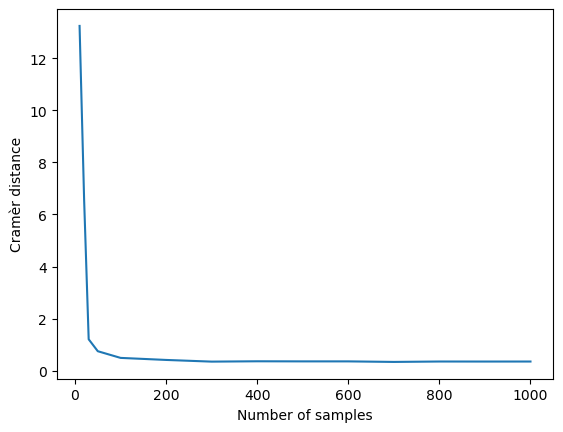}
        \caption{Impact of the number of base distribution samples on performance. 
        As the number of samples used for computing the distributional loss increases, 
        NFDRL's loss decreases and performance improves. 
        A plateau is reached around 100 samples, indicating a good trade-off between accuracy and efficiency.}
    \end{subfigure}
    \caption{NFDRL Parameter Efficiency.}
    \label{fig:n_params}
\end{figure}

We evaluate parameter efficiency by comparing our model to C51. Unlike C51, whose parameter count grows with the number of atoms, our model maintains a constant size and already matches C51 with just 11 atoms, highlighting its superior efficiency (Figure \ref{fig:n_params} (left)). We further examine the impact of the number of base distribution samples in MDP2, observing that performance improves as the sample count increases and stabilizes around 100 samples, which offers a good balance between accuracy and computational cost (Figure \ref{fig:n_params} (right)).

\section{Conclusion}

We introduced a new DistRL method that models return distributions as mixtures of Gaussians, with parameters learned via normalizing flows. By optimizing the Cramér loss—exactly or through a surrogate—we capture richer uncertainty and learn more precise value distributions. Empirically, our method achieves competitive or better performance on Atari games while being far more parameter-efficient than C51 and more expressive than quantile-based methods.

\textbf{Limitations.} Nonetheless, our approach bears certain limitations. While mentioned here, these are more detailed in the appendix section 1. (i) Our method shows high training variance, especially on certain games, partly due to sensitivity to learning rate schedules and stochasticity from sampling. (ii) Training convergence is relatively slow, stemming from variance, the indirect nature of flow parameterization, and the inherent cost of training normalizing flows. (iii) Design choices such as using CDF flows and KDE for target distributions introduce modeling constraints, inefficiencies, and non-trainable components.


\bibliography{Bibliography} 

\newpage
\appendix


\setcounter{figure}{0}             
\renewcommand{\thefigure}{A\arabic{figure}} 


\title{Parameter-Efficient Distributional RL via Normalizing Flows and a Geometry-Aware Cramèr Surrogate}

\section{Limitations}

\begin{figure}[h!]
    \centering
    \includegraphics[width=0.5\linewidth]{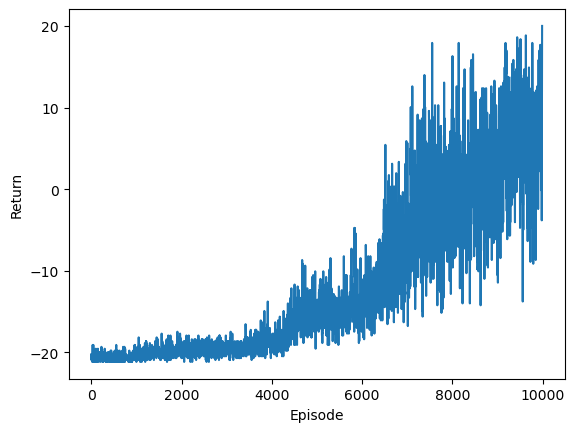}
    \caption{Training curve on PONG without learning rate decay}
    \label{fig:Pong_return}
\end{figure}

\paragraph{Variance} Our method exhibits high training variance, particularly on games such as Pong. This sensitivity can be partially mitigated by carefully tuning the learning rate schedule, a hyperparameter to which our model is notably sensitive. Figure~\ref{fig:Pong_return} illustrates a training run using a fixed learning rate: while the model eventually achieves high performance, its learning curve remains highly unstable. Furthermore, because our loss function relies on sampling from base distributions (e.g., for computing the Cramér distance), it introduces additional stochasticity that contributes to this variance.

\paragraph{Learning efficiency} As shown in main paper table 1 and Appendix figure \ref{fig:Pong_return}, our model achieves competitive performance but at the cost of a slow training convergence. We believe this can be due to different factors:
\begin{enumerate}
    \item The training variance might not help the model converge faster
    \item Instead of learning directly the density of given values like C51, or specified values, our model learns flow parametrisations that indirectly lead to return distributions. This indirect relationship might hinder the learning performance by making the task more complex for the model.
    \item Normalizing Flows are effective for learning exact likelihoods but they are notoriously slow to train, this fact is confirmed by our empirical results.
\end{enumerate} 

\paragraph{CDF Flow} While using a CDF as a flow transformation offers advantages in modeling monotonic mappings and enabling efficient computation of the Cramér distance (Main paper section 3.2), it also introduces notable limitations. First, the CDF is inherently bounded, making it ill-suited for modeling unbounded return distributions without an additional projection step to extend its support. Second, since the model parameterizes a mixture of Gaussians, the components can become disjoint—e.g., with widely separated means and variances—which may result in poor overlap with the base distribution (typically standard Gaussian). Consequently, this can lead to inefficient coverage of the learned CDF's support, introducing instability and training inconsistencies. Although chaining multiple flow transformations or adopting alternative flow families may alleviate this issue, doing so increases model complexity and may hinder convergence. Addressing this trade-off remains an open direction for future work.

\paragraph{KDE for Target Distribution} 
To enable the computation of the Cramér distance, we construct a target return distribution via KDE, ensuring that it shares the same support as the predicted distribution. However, this introduces a non-trivial computational burden and a non-trainable step in the pipeline, as gradients do not flow through the KDE. We attempted to mitigate this by reusing the same set of samples $z$ (used to parameterize the predicted distribution) to build the KDE target, in the hope of improving alignment and efficiency. Unfortunately, this strategy did not yield significant improvements. Ideally, a more integrated approach would avoid the need for KDE altogether, enabling fully end-to-end training and reducing reliance on handcrafted alignment procedures.

\section{Alignment Procedure and Algorithm}
As mentionned in main paper section 3.4, our method necessitates a support alignment between the predict and target distributions. Indeed for the same samples $z_i\sim\mathcal{U}$, we have $F_{(x,a)}(z_i)=y_i$ and the target distribution is based on subsequent states and actions, therefore $\tilde{y}_i=F_{(x',a')}(z)$. Said otherwise, as the predicted and target distributions are based using different state-action pairs, the output return values are different are also different, without even taking the bootstrap function into account. Therefore, in order to compare the two distributions accurately, we use a KDE to get $\eta^\pi(x,a)(y_i), \eta^\pi(x,a)(\tilde{y}_i), T^\pi\eta(x',a')(y_i), T^\pi\eta(x',a')(\tilde{y}_i)$. 

Let $y_i \sim \eta^\pi(x,a)$ be samples from the predicted return distribution and $\tilde{y}_j \sim T^\pi \eta(x,a)$ be samples from the target distribution. First, we evaluate the KDE of $T^\pi \eta(x,a)$ on the predicted support:
\[T^\pi \eta(x, a)(y_i) = \frac{1}{M} \sum_{j=1}^M K_h(y_i - \tilde{y}_j), \quad \forall i = 1, \dots, N,\]
and reciprocally:
\[\hat{\eta}^\pi(x, a)(\tilde{y}_j) = \frac{1}{N} \sum_{i=1}^N K_h(\tilde{y}_j - y_i), \quad \forall j = 1, \dots, M.\]

\begin{figure}[h!]
    \centering
    \includegraphics[width=\textwidth]{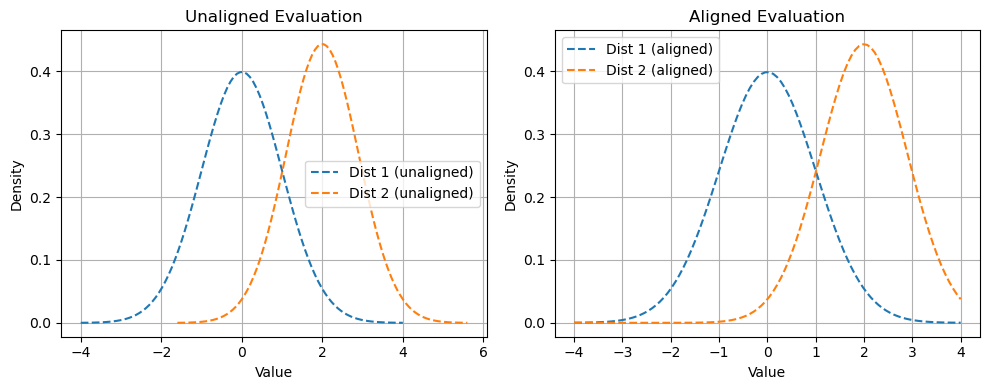}
    \caption{ Impact of Support Alignment on Distribution Comparison. Left: Two Gaussian distributions evaluated on different support sets, making direct comparison ill-posed. Right: The same distributions evaluated on a shared support, enabling meaningful density-wise comparison. This illustrates the necessity of support alignment in distributional reinforcement learning, akin to the projection step in C51.}
    \label{fig:alignment}
\end{figure}

The process is illustrated in figure \ref{fig:alignment} and detailed in algorithm \ref{alg:target}.

\begin{algorithm}[H] 
\caption{Flow-based Target Distribution Construction in Distributional RL}
\begin{algorithmic}
\Require Current return distribution $\eta^\pi(x, a)$, reward $r$, next state $x'$, terminal indicator $d$, support $z$, number of samples $N$

\State Compute next-state return distribution:
\[
(y', p^\pi(y')) \gets g(F_\theta(x', z), z)
\]

\If{$x'$ is terminal}
    \State Replace $(y', p^\pi(y'))$ with Gaussian $\mathcal{N}(r, 0.1)$
\EndIf

\State Compute expected returns:
\[
Q^\pi(x', a') = \sum_{y'} y' \cdot  p^\pi(y'))
\]

\State Greedy action: $a^* = \arg\max_{a'} Q^\pi(x', a')$
\State Select corresponding distribution: $y^*=y^*[a^*]$
\State Select support $s=[-\max(y,y^*),\max(y,y^*)]$

\State Estimate KDEs:
\[
\hat{p}_{\eta}(s) \gets \text{KDE}(y,\eta(x,a)(y)), \quad \hat{p}_{T^\pi\eta}(s) \gets \text{KDE}(y^*,  T^\pi\eta(x',a^*)(y^*))
\]

\State Interpolate both to support $\{y_i\}_{i=1}^N$:
\[
p_{\eta}(y_i), p_{T^\pi\eta}(y_i) \gets \text{Interpolate}(\hat{p}_{\eta}, \hat{p}_{T^\pi\eta})
\]

\State Interpolate both to support $\{\tilde{y}_j\}_{j=1}^N$:
\[
\tilde{p}_{\eta}(\tilde{y}_j), \tilde{p}_{T^\pi\eta}(\tilde{y}_j) \gets \text{Interpolate}(\hat{p}_{\eta}, \hat{p}_{T^\pi\eta})
\]

\State \Return $\{\tilde{y}_j\}_{j=1}^N, p_{\eta}(y_i), p_{T^\pi\eta}(y_i), \tilde{p}_{\eta}(\tilde{y}_j), \tilde{p}_{T^\pi\eta}(\tilde{y}_j)$ 

\end{algorithmic}
\label{alg:target}
\end{algorithm}

\section{Toy Markov decision processes}

\begin{figure}[h!]
    \centering
   \begin{subfigure}{0.45\textwidth}
        \centering
        \begin{tikzpicture}[->, >=Stealth, node distance=2cm]

    \tikzstyle{state}=[circle, draw, minimum size=1.2cm]
    \tikzstyle{reward}=[rectangle, draw, minimum size=1cm]

    \node[state] (s1) {$s_1$};
    \node[state] (s2) [right of=s1] {$s_2$};
    \node[reward] (R1) [above of=s2] {$R_1$};
    \node[state] (s3) [right of=s2] {$s_3$};
    \node[reward] (R2) [above of=s3] {$R_2$};

    \draw (s1) -- node[above] {$a_1$} (s2);
    \draw[dotted] (s2) -- (R1);
    \draw (s2) -- node[above] {$a_1$} (s3);
    \draw[dotted] (s3) -- (R2);

    \end{tikzpicture}
    \caption*{$MDP_1$}
     \end{subfigure}
    \hfill
    \begin{subfigure}{0.45\textwidth}
        \centering

    \begin{tikzpicture}[->, >=Stealth, node distance=2cm]

    \tikzstyle{state}=[circle, draw, minimum size=1.2cm]
    \tikzstyle{reward}=[rectangle, draw, minimum size=1cm]

    \node[state] (s1) {$s_1$};
    \node[state] (s2) [above right of=s1] {$s_2$};
    \node[state] (s3) [below right of=s1] {$s_3$};
    \node[reward] (R1) [right of=s2] {$R_1$};
    \node[reward] (R2) [right of=s3] {$R_3$};

    \draw (s1) -- node[above left] {$a_1,p=0.5$} (s2);
    \draw (s1) -- node[below left] {$a_1,p=0.5$} (s3);
    \draw[dotted] (s2) -- (R1);
    \draw[dotted] (s3) -- (R2);

\end{tikzpicture}
\caption*{$MDP_2$}
     \end{subfigure}
 \caption{Two example MDPs.}
\label{fig:simpleMDP}
\end{figure}

\begin{figure}[h]
    \centering
\begin{tikzpicture}[->, >=Stealth, node distance=2cm]

    \tikzstyle{state}=[circle, draw, minimum size=1.2cm]
    \tikzstyle{reward}=[rectangle, draw, minimum size=1cm]

    \node[state] (s1) {$s_1$};
    \node[state] (s2) [right of=s1] {$s_2$};
    \node[reward] (R1) [above of=s2] {$R_1=0$};
    \node[state] (s3) [right of=s2] {$s_3$};
    \node[reward] (R2) [above of=s3] {$R_2=0$};
    \node[state] (s4) [right of=s3] {$s_4$};
    \node[reward] (R3) [above right=1cm of s4] {$R_3 \sim \mathcal{N}(-2,1)$};
    \node[reward] (R4) [below right=1cm of s4] {$R_4 \sim \mathcal{N}(2,0.8)$};

    \draw (s1) -- node[above] {$a_1$} (s2);
    \draw[dotted] (s2) -- (R1);
    \draw (s2) -- node[above] {$a_1$} (s3);
    \draw (s3) -- node[above] {$a_1$} (s4);
    \draw[dotted] (s3) -- (R2);
    \draw[dotted] (s4) -- node[above] {$0.5$} (R3);
    \draw[dotted] (s4) -- node[above] {$0.5$} (R4);
    \end{tikzpicture}
    \caption{$MDP_3; \gamma =1$}  
\label{fig:MDP3}
\end{figure}

\begin{figure}[h!]
    \centering
    \begin{subfigure}{0.3\textwidth}
        \includegraphics[width=\linewidth]{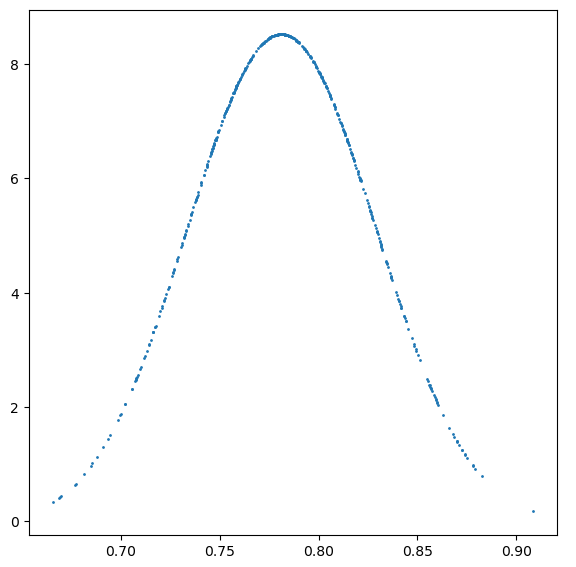}
        \caption{bandwith: $0.05$, s.e: $0.1$}
    \end{subfigure}
    \hfill
    \begin{subfigure}{0.3\textwidth}
        \includegraphics[width=\linewidth]{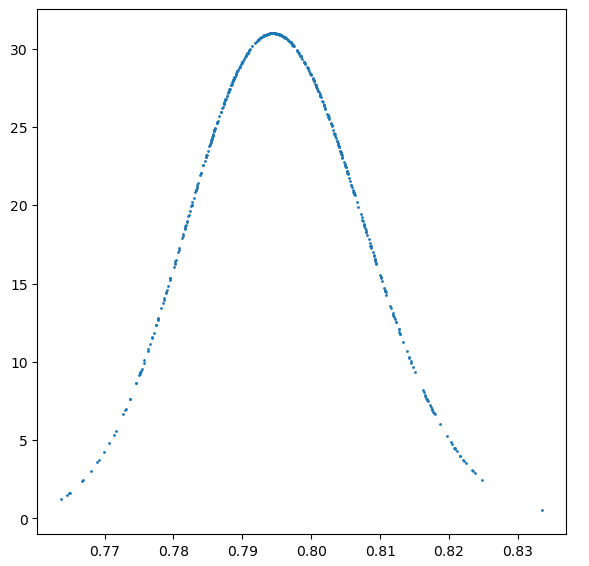}
        \caption{bandwith: $0.01$, s.e: $0.01$}
    \end{subfigure}
    \hfill
    \begin{subfigure}{0.3\textwidth}
        \includegraphics[width=\linewidth]{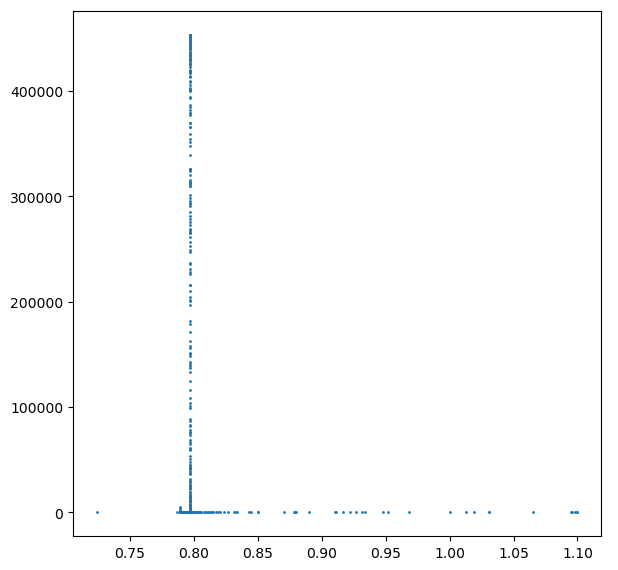}
        \caption{bandwith: $0.001$, s.e: $0.01$}
    \end{subfigure}
    
    \vskip\baselineskip 

    \begin{subfigure}{0.3\textwidth}
        \includegraphics[width=\linewidth]{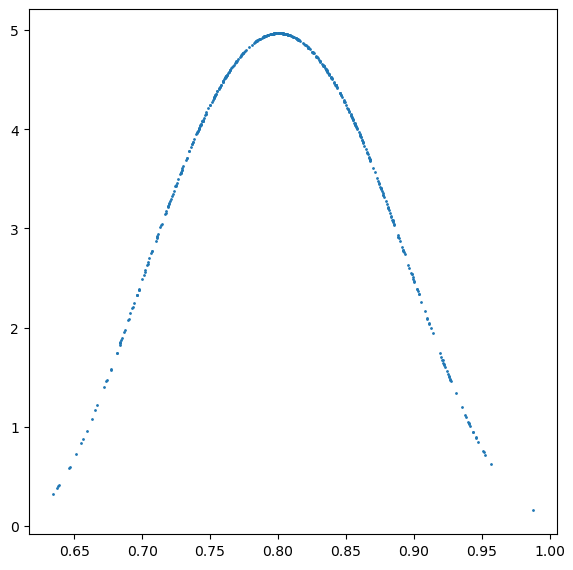}
        \caption{bandwith: $0.05$, s.e: $0.1$}
    \end{subfigure}
    \hfill
    \begin{subfigure}{0.3\textwidth}
        \includegraphics[width=\linewidth]{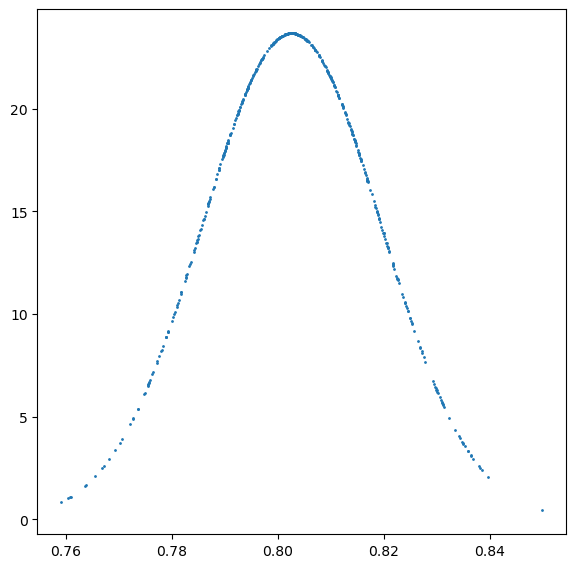}
        \caption{bandwith: $0.01$, s.e: $0.01$}
    \end{subfigure}
    \hfill
    \begin{subfigure}{0.3\textwidth}
        \includegraphics[width=\linewidth]{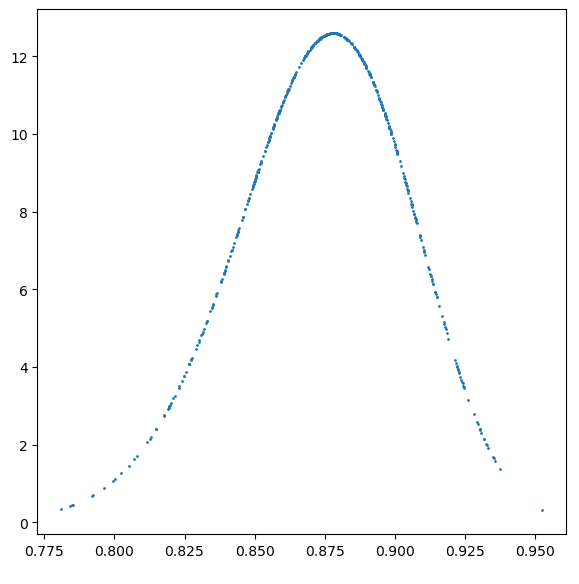}
        \caption{bandwith: $0.001$, s.e: $0.01$}
    \end{subfigure}

    \caption{Learnt return distributions for the state-action pair $(s_2, a_1)$ in $MDP_1$, under different values of the KDE bandwidth and final state's reward variance. The x-axis is return values and the y axis corresponds to their corresponding densities. The target reward is $0.8$. (a,b,c) show distributions learned using the exact Cramér loss, while (d,e,f) show those obtained with our surrogate. Narrower distributions can be achieved with both losses, illustrating the method's flexibility.}
    \label{fig:final}
\end{figure}

\begin{figure}[h!]
    \centering
    \begin{subfigure}{0.45\textwidth}
    \includegraphics[width=\linewidth]{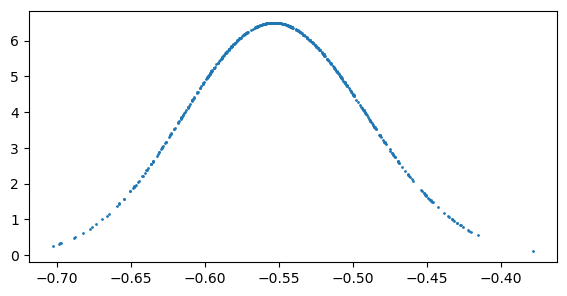}
        \caption{Exact Cramér}
    \end{subfigure}
    \hfill
    \begin{subfigure}{0.45\textwidth}
    \includegraphics[width=\linewidth]{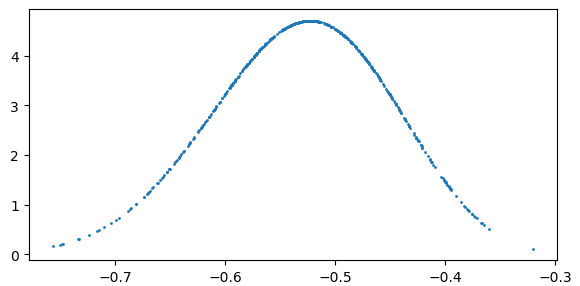}
        \caption{Surrogate}
    \end{subfigure}
    \caption{Learnt distributions for $(s_1,a_1)$ in $MDP_1$; $R_1=-0.8$, $R_2=0.3$; $\gamma=0.9$. 
    The x-axis is return values and the y axis corresponds to their densities. 
    A KDE bandwidth of $0.05$ and a final reward standard error of $0.01$ are used.}
    \label{fig:MDP}
\end{figure}

\begin{figure}[h!]
    \centering
    \begin{subfigure}{0.3\textwidth}
    \includegraphics[width=\linewidth]{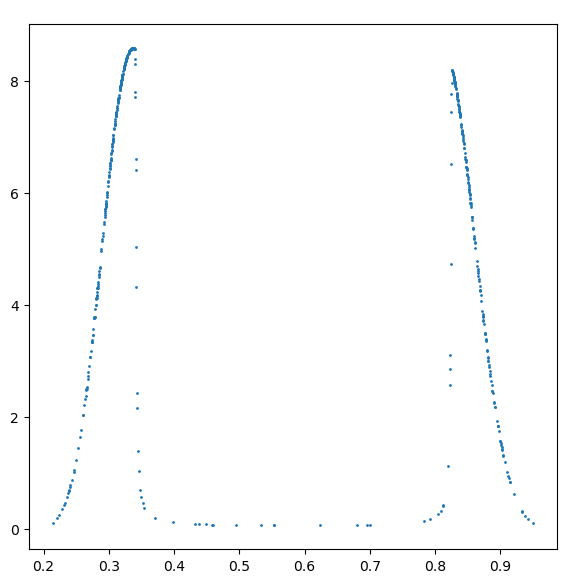}
        \caption{Exact Cramér}
    \end{subfigure}
    \hfill
    \begin{subfigure}{0.3\textwidth}
    \includegraphics[width=\linewidth]{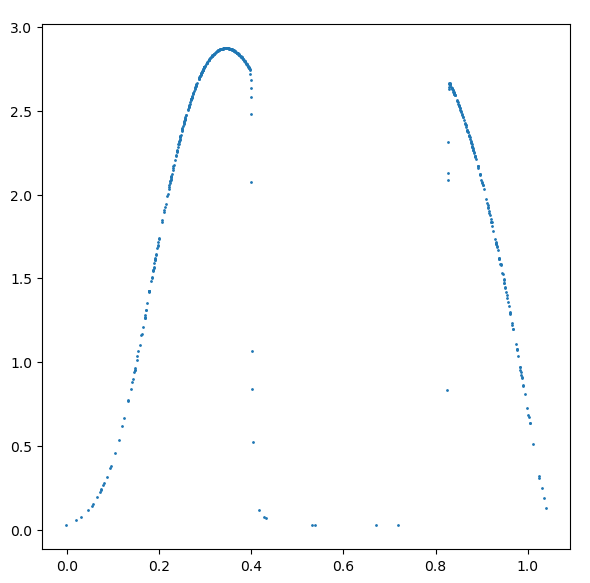}
        \caption{Surrogate}
    \end{subfigure}
    \hfill
    \begin{subfigure}{0.3\textwidth}
    \includegraphics[width=\linewidth]{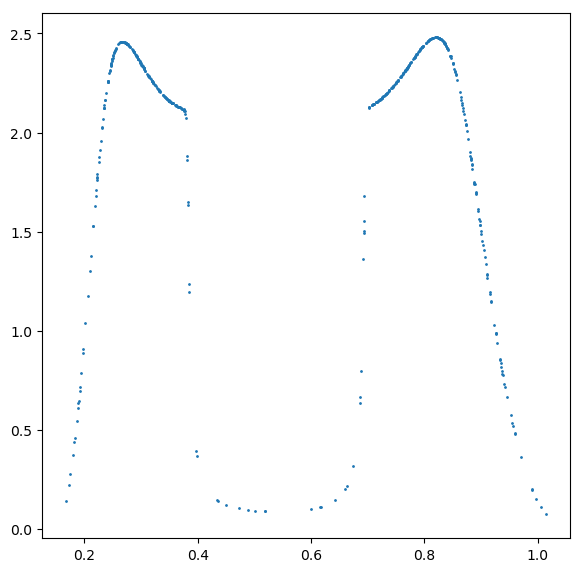}
        \caption{$L^2$ loss}
    \end{subfigure}
    \caption{Learnt distributions for $(s_1,a_1)$ in $MDP_2$; $R_1=0.8$, $R_2=0.3$. 
    The x-axis is return values and the y-axis corresponds to their corresponding densities. 
    We use a KDE bandwidth of $0.05$ and a final reward standard error of $0.1$.}
    \label{fig:bimodal}
\end{figure}

\clearpage

\section{Relevant distance properties}

\citet{bellemare2017cramer} show that the Cramér distance holds the following properties:

\textbf{Scale sensitivity}. Consider a divergence $\mathbf{d}$, and for two random variables $X,Y$ with distributions $P,Q$, write $\mathbf{d}(X,Y):= \mathbf{d}(P,Q)$. We say that $\mathbf{d}$ is scale sensitive (of order $\beta$), i.e. it has property ($\mathbf{S})$ if there exists a $\beta>0$ such that for all $X,Y$, and a real value $c>0$,
\begin{equation}\tag{$\mathbf{S}$}
\mathbf{d}(cX,cY)\leq|c|^\beta\mathbf{d}(X,Y)
\end{equation}

\textbf{Sum invariance}: A divergence $\mathbf{d}$ has property $\mathbf{(I)}$, i.e. it is sum invariant, if whenever $A$ is independent from $X,Y$
\begin{equation}\tag{$\mathbf{I}$}
\mathbf{d}(A+X,A+Y)\leq\mathbf{d}(X,Y)
\end{equation}

A divergence is said ideal if it possesses both $\mathbf{(S)}$ and $\mathbf{(I)}$. 

\textbf{Unbiased sample gradients}: Let $X_1,X_2,\dots,X_m$ be independent samples from $P$ and define the empirical distribution $\hat{P}_m:= \hat{P}_m(X_m):=\frac{1}{m}\sum_{i=1}^m\delta_{X_i}$. From this, define the sample loss $\mathbf{d}(\hat{P}_m,Q_\theta)$. We say that $\mathbf{d}$ has unbiased sample gradients when the expected gradient of the sample loss equals the gradient of the true loss for all $P$ and $m$:
\begin{equation}\tag{$\mathbf{U}$}
\mathbb E_{X_m\sim P}\nabla_\theta\mathbf{d}(\hat{P}_m,Q_\theta)=\nabla_\theta\mathbf{d}(P,Q_\theta)
\end{equation}

If a divergence does not possess $\mathbf{(U)}$, then minimising it with stochastic gradient descent may not converge or towards the wrong minimum. Conversely, if $\mathbf{d}$ possesses $\mathbf{(U)}$ then we can guarantee that the distribution which minimises the expected sample loss is $Q = P$. From these properties, \cite{bellemare2017cramer} draws the following propositions:\\
\textbf{Proposition 1:} \textit{The KL divergence has unbiased sample gradients (U), but is not scale sensitive (S).}\\
\textbf{Proposition 2:} \textit{The Wasserstein metric is ideal (I, S), but does not have unbiased sample gradients.}

\section{Cramér-inspired geometry-aware metric}\label{sec:cramer}

We introduce a Cramér-inspired, geometry-aware metric on discrete probability masses and show that it enjoys key properties (metric axioms, $\sqrt{gamma}$-contraction, and unbiased sample gradients) mirroring those that make the Cramér distance appealing in DistRL. Recall that for two distributions with CDFs $F(x)$ and $G(x)$, the squared Cramér distance is:
\begin{equation*}
    l_2^2(P,Q) = \int_{-\infty}^{\infty} (F(x)-G(x))^2 \, dx
\end{equation*}

Let $\delta(t) = p(t) - q(t)$ be the difference in probability densities. We can express the CDF difference as an integral:
\begin{equation*}
    F(x)-G(x) = \int_{-\infty}^{x} \delta(t) \, dt
\end{equation*}

Substituting this into the distance definition:
\begin{equation*}
    l_2^2(P,Q) = \int_{-\infty}^{\infty} \left( \int_{-\infty}^{x} \delta(t) \, dt \right)^2 \, dx
\end{equation*}

We apply the Cauchy-Schwarz inequality to bound the inner term. For any interval $[y,x]$, we have:
\begin{equation*}
    \left( \int_{y}^{x} \delta(t) \cdot 1 \, dt \right)^2 \le \left( \int_{y}^{x} 1^2 \, dt \right) \left( \int_{y}^{x} \delta(t)^2 \, dt \right)
\end{equation*}

The first term on the right side evaluates simply to the length of the interval:
\begin{equation*}
    \int_{y}^{x} 1 \, dt = |x-y|
\end{equation*}

Thus, we obtain the upper bound:
\begin{equation*}
    (F(x)-G(x))^2 \le |x-y| \int_{-\infty}^{x} \delta(t)^2 \, dt
\end{equation*}

Substituting this back into the outer integral, and swapping the order of integration (Fubini’s theorem), we approximate the global distance as a pairwise sum over the support. Discretizing the domain into $N$ support points $\{y_1, \dots, y_N\}$ with associated probability masses $w_i$ and $v_i$ (where $\delta(y_i) \approx w_i - v_i$), the integral approximates to:
\begin{equation*}
    \mathcal{L}_{\text{surrogate}} = \sum_{i=1}^{N} \sum_{j=1}^{N} (w_i - v_i)^2 \cdot |y_i - y_j|
\end{equation*}

In the next sections we will investigate the three following questions:
\begin{itemize}
    \item \textbf{Q1:} Is this loss a proper distance? If so, then the loss function would be symmetric and scale sensitive unlike KL divergence.
    \item \textbf{Q2:} Is the distributional Bellman operator a contraction in this case? If so, this would ensure convergence of the distributional Bellman operator $\eta^\pi$ towards the random returns $T^\pi_\eta$.
    \item \textbf{Q3:} Does it still possess the unbiased sample gradient estimate property? If so, then SGD can be used to optimise this loss function in a RL context unlike the Wasserstein distance. More specifically, we will be able to learn from sample transitions.
\end{itemize}

In what follows, the theory treats the masses $w_i$ abstractly; they may arise from a Riemann approximation of a continuous density or from an empirical distribution over Monte-Carlo samples. Our proofs only require that they form valid discrete probability distributions.

\subsection{Q1: Is it a Proper Distance?}

We verify that the proposed loss function satisfies the four axioms of a metric: Non-negativity, Symmetry, Identity of Indiscernibles, and the Triangle Inequality.

Let the loss be defined over the discrete probability masses $\mathbf{w}, \mathbf{v} \in \mathbb{R}^N$ defined on a fixed support $\{y_1, \dots, y_N\}$:
\begin{equation}
    D(\mathbf{w}, \mathbf{v}) = \sqrt{\frac{1}{N^2} \sum_{i=1}^N \sum_{j=1}^N (w_i - v_i)^2 |y_i - y_j|}
\end{equation}

First, we simplify the expression by factoring out the term that depends only on $i$. Let $\Omega_i$ be the "geometric weight" of the $i$-th support point, representing the sum of distances from $y_i$ to all other points:
\begin{equation}
    \Omega_i = \sum_{j=1}^N |y_i - y_j|
\end{equation}
Note that as long as $N > 1$ and the support points are distinct, $\Omega_i > 0$ for all $i$.

Substituting this into the loss equation:
\begin{equation}
    D(\mathbf{w}, \mathbf{v}) = \frac{1}{N} \sqrt{\sum_{i=1}^N \Omega_i (w_i - v_i)^2}
\end{equation}
This reveals that our loss function is simply a Weighted Euclidean Distance (scaled by $1/N$) between the mass vectors $\mathbf{w}$ and $\mathbf{v}$, with weights $\Omega_i$.

\begin{enumerate}[leftmargin=*]
    \item \textbf{Non-negativity:}
    Since $(w_i - v_i)^2 \ge 0$ and weights $\Omega_i > 0$, the sum is non-negative. The square root function preserves non-negativity. Thus, $D(\mathbf{w}, \mathbf{v}) \ge 0$.

    \item \textbf{Symmetry:}
    The term $(w_i - v_i)^2$ is inherently symmetric: $(w_i - v_i)^2 = (v_i - w_i)^2$. The weights $\Omega_i$ depend only on the support geometry, which is fixed for the comparison. Thus, $D(\mathbf{w}, \mathbf{v}) = D(\mathbf{v}, \mathbf{w})$.

    \item \textbf{Identity of Indiscernibles:}
    Clearly, if $\mathbf{w} = \mathbf{v}$, then $(w_i - v_i) = 0$ for all $i$, so $D=0$.
    Conversely, assume $D(\mathbf{w}, \mathbf{v}) = 0$. This implies:
    \begin{equation*}
        \sum_{i=1}^N \Omega_i (w_i - v_i)^2 = 0
    \end{equation*}
    Since $\Omega_i > 0$ strictly (for distinct support points with $N > 1$) and $(w_i - v_i)^2 \ge 0$, the only way the sum can be zero is if every individual term $(w_i - v_i)^2 = 0$. Therefore, $w_i = v_i$ for all $i$, implying $\mathbf{w} = \mathbf{v}$.

    \item \textbf{Triangle Inequality:}
    We must show that $D(\mathbf{w}, \mathbf{u}) \le D(\mathbf{w}, \mathbf{v}) + D(\mathbf{v}, \mathbf{u})$ for any third distribution $\mathbf{u}$.

    Let us define the transformed vector $\mathbf{x}$ such that $x_i = \sqrt{\Omega_i} w_i$. Similarly, let $x'_i = \sqrt{\Omega_i} v_i$ and $x''_i = \sqrt{\Omega_i} u_i$.

    The loss can be rewritten as the standard Euclidean ($l_2$) distance between these transformed vectors (ignoring the $1/N$ factor, which scales all terms linearly):
    \begin{equation*}
        N \cdot D(\mathbf{w}, \mathbf{v}) = \sqrt{\sum_{i=1}^N (\sqrt{\Omega_i} w_i - \sqrt{\Omega_i} v_i)^2} = ||\mathbf{x} - \mathbf{x}'||_2
    \end{equation*}

    The standard Euclidean norm satisfies the triangle inequality:
    \begin{equation*}
        ||\mathbf{x} - \mathbf{x}''||_2 \le ||\mathbf{x} - \mathbf{x}'||_2 + ||\mathbf{x}' - \mathbf{x}''||_2
    \end{equation*}

    Substituting back the definitions, we obtain:
    \begin{equation*}
        N \cdot D(\mathbf{w}, \mathbf{u}) \le N \cdot D(\mathbf{w}, \mathbf{v}) + N \cdot D(\mathbf{v}, \mathbf{u})
    \end{equation*}

    Dividing by $N$, the triangle inequality holds.
\end{enumerate}

\paragraph{Conclusion:}
The proposed surrogate loss $D$ defines a valid metric on the space of discrete probability distributions over a fixed support.

\subsection{Q2: Is the Distributional Bellman Operator a Contraction for D?}

We now show that the distributional Bellman operator $\mathcal{T}^\pi$ is a contraction under the metric $D$ defined above.

Recall the definition of the operator. For a transition $(x, a, r, x')$, the operator applies a shift and scale to the random return $Z(x', a')$. If $Z \sim \eta$, then the target return is $R + \gamma Z$.

In our discrete mass framework, this transformation affects the support locations $\{y_i\}$, while leaving the probability masses invariant (as established in the implementation note).

Let $\mathbf{w}$ and $\mathbf{v}$ be the probability mass vectors for two different return distributions $\eta_1$ and $\eta_2$ defined on the same support $\{y_i\}$.
The distance between them is:
\begin{equation}
    D(\eta_1, \eta_2) = \sqrt{\frac{1}{N^2} \sum_{i,j} (w_i - v_i)^2 |y_i - y_j|}
\end{equation}

Applying the Bellman operator $\mathcal{T}^\pi$ transforms the support points $y_i$ to new locations $y'_i = r + \gamma y_i$. As discussed, the probability masses $w_i$ and $v_i$ associated with these points remain invariant under this transformation.

Thus, the distance between the projected distributions is:
\begin{equation}
    D(\mathcal{T}^\pi \eta_1, \mathcal{T}^\pi \eta_2) = \sqrt{\frac{1}{N^2} \sum_{i,j} (w_i - v_i)^2 |y'_i - y'_j|}
\end{equation}

Substituting $y'_i = r + \gamma y_i$:
\begin{equation}
    |y'_i - y'_j| = |(r + \gamma y_i) - (r + \gamma y_j)| = |\gamma(y_i - y_j)| = \gamma |y_i - y_j|
\end{equation}
(Note: Since $\gamma > 0$, $|\gamma| = \gamma$).

Substituting this back into the distance equation:
\begin{equation}
\begin{aligned}
    D(\mathcal{T}^\pi \eta_1, \mathcal{T}^\pi \eta_2) &= \sqrt{\frac{1}{N^2} \sum_{i,j} (w_i - v_i)^2 \cdot \gamma |y_i - y_j|} \\
    &= \sqrt{\gamma} \cdot \sqrt{\frac{1}{N^2} \sum_{i,j} (w_i - v_i)^2 |y_i - y_j|} \\
    &= \sqrt{\gamma} \cdot D(\eta_1, \eta_2)
\end{aligned}
\end{equation}

\paragraph{Conclusion:}
Since the discount factor satisfies $0 < \gamma < 1$, it follows that $\sqrt{\gamma} < 1$. Therefore, the distributional Bellman operator $\mathcal{T}^\pi$ is a contraction mapping with respect to the metric $D$, with contraction factor $\sqrt{\gamma}$.

By Banach's Fixed Point Theorem, repeated application of $\mathcal{T}^\pi$ converges to a unique fixed-point distribution.

\subsection{Q3: Does d Admit Unbiased Sample Gradient Estimates?}\label{sec:Q3}

Finally, we verify that our proposed surrogate loss satisfies property (U) (Unbiased Sample Gradients) as defined by \cite{bellemare2017cramer}. This property ensures that minimizing the loss over stochastic samples (e.g., transitions sampled from the replay buffer) minimizes the true expected loss over the full distribution.

Let $\mathbf{w}(\theta)$ be the predicted probability mass vector parameterized by $\theta$. Let $\mathbf{v}_{\text{true}}$ be the true target probability mass vector. In a reinforcement learning setting, we do not observe $\mathbf{v}_{\text{true}}$ directly; instead, we observe empirical samples (realizations of returns) which form an empirical target distribution $\hat{\mathbf{v}}$.

We define the squared sample loss $\mathcal{L}$ as:
\begin{equation}
    \mathcal{L}(\theta, \hat{\mathbf{v}}) = \sum_{i=1}^N \sum_{j=1}^N (w_i(\theta) - \hat{v}_i)^2 |y_i - y_j|
\end{equation}

To simplify notation, let $\Omega_i = \sum_{j=1}^N |y_i - y_j|$ be the geometric weight associated with support point $i$. The loss simplifies to a weighted sum of squared errors:
\begin{equation}
    \mathcal{L}(\theta, \hat{\mathbf{v}}) = \sum_{i=1}^N \Omega_i (w_i(\theta) - \hat{v}_i)^2
\end{equation}

We compute the gradient with respect to parameters $\theta$:
\begin{equation}
    \nabla_\theta \mathcal{L} = \sum_{i=1}^N 2 \Omega_i (w_i(\theta) - \hat{v}_i) \nabla_\theta w_i(\theta)
\end{equation}

We now examine the expected gradient over the distribution of sample targets. Since the expectation is a linear operator:
\begin{equation}
\begin{aligned}
    \mathbb{E}_{\hat{\mathbf{v}}} [\nabla_\theta \mathcal{L}] &= \mathbb{E}_{\hat{\mathbf{v}}} \left[ \sum_{i=1}^N 2 \Omega_i (w_i(\theta) - \hat{v}_i) \nabla_\theta w_i(\theta) \right] \\
    &= \sum_{i=1}^N 2 \Omega_i (w_i(\theta) - \mathbb{E}[\hat{v}_i]) \nabla_\theta w_i(\theta)
\end{aligned}
\end{equation}

Since $\hat{\mathbf{v}}$ is constructed from unbiased samples of the return (e.g., via Monte Carlo sampling or an unbiased empirical distribution), we have $\mathbb{E}[\hat{v}_i] = v_{\text{true},i}$. Therefore:
\begin{equation}
    \mathbb{E}_{\hat{\mathbf{v}}} [\nabla_\theta \mathcal{L}] = \nabla_\theta \mathcal{L}(\theta, \mathbf{v}_{\text{true}})
\end{equation}

\paragraph{Conclusion:}
The expected gradient of the sample loss is exactly the gradient of the true loss. Unlike the Wasserstein distance, which generally requires biased sample gradients or dual approximations, our geometry-aware surrogate allows for direct stochastic gradient descent (SGD) on sample transitions.

\section{Behaviour of the Geometry-Aware Metric and the Role of KDE}\label{KDE_appendix}

In this section we clarify a limitation of the discrete geometry-aware metric $D$ in a simple ``one-hot, disjoint support'' setting, and then show how the KDE-based construction used in practice mitigates this issue and yields a loss that behaves more like a transport distance.

\subsection{A limitation of the exact metric for disjoint one-hot distributions}

Recall that for a fixed support $\{y_1,\dots,y_N\}$ and two discrete probability mass vectors
\[
w = (w_1,\dots,w_N), \qquad v = (v_1,\dots,v_N),
\]
our surrogate Cramér distance is
\begin{equation}
    D(w,v) 
    \;=\; 
    \sqrt{\frac{1}{N^2} \sum_{i=1}^N \sum_{j=1}^N (w_i - v_i)^2 \, |y_i - y_j|}
    \;=\;
    \frac{1}{N}\sqrt{\sum_{i=1}^N \Omega_i (w_i - v_i)^2},
    \label{eq:geom_D_app}
\end{equation}
with
\begin{equation}
    \Omega_i \;:=\; \sum_{j=1}^N |y_i - y_j|.
    \label{eq:geom_Omega_app}
\end{equation}

Consider now the following simple discrete setting:
\begin{itemize}
    \item the support $\{y_i\}$ is a fixed grid (e.g., uniform on $[-K,K]$);
    \item $w$ and $v$ are one-hot distributions:
    \[
    w_i = \mathbbm{1}\{i=a\}, \qquad v_i = \mathbbm{1}\{i=b\}
    \]
    for some indices $a \neq b$.
\end{itemize}
In this case,
\[
(w_i - v_i)^2 = 
\begin{cases} 
1, & i \in \{a,b\}, \\
0, & \text{otherwise},
\end{cases}
\]
and therefore the double sum in \eqref{eq:geom_D_app} collapses to
\begin{equation}
    D^2(w,v) = \frac{1}{N^2} \big(\Omega_a + \Omega_b\big).
    \label{eq:geom_D_onehot}
\end{equation}
A key observation is that $D^2(w,v)$ in \eqref{eq:geom_D_onehot} does not depend explicitly on the distance $|y_a - y_b|$ between the two spikes, except through which weights $\Omega_a$ and $\Omega_b$ happen to be selected. On a symmetric uniform grid, $\Omega_i$ is smallest in the centre and largest near the edges, but for two disjoint one-hot distributions the value of $D(w,v)$ is determined by:
\begin{enumerate}
    \item which bins carry mass (via $\Omega_a, \Omega_b$),
    \item not by how far apart those bins are.
\end{enumerate}
Concretely, on a grid $\{-10, -9, \dots, 10\}$, one can compute
\[
\Omega_{-10} = \Omega_{+10} = 210, \qquad \Omega_{-9} = 191,
\]
so that
\[
D^2(\delta_{-10}, \delta_{-9}) \;\propto\; 210 + 191 = 401, 
\qquad
D^2(\delta_{-10}, \delta_{+10}) \;\propto\; 210 + 210 = 420,
\]
which are very close despite the spikes being at distance $1$ vs $20$. In other words:

For disjoint one-hot distributions, the exact metric $D$ behaves as a geometry-weighted Euclidean distance on bin masses, and does not strongly distinguish ``near'' from ``far'' spikes in the way an optimal-transport or Cramér distance does.

This is the core limitation that will be addressed by the KDE-based construction used in our practical loss.

\subsection{KDE-based construction of the practical loss}

In the model, the return distributions are represented as continuous densities via normalizing flows. For a given $(x,a)$, denote
\[
p(y) := \eta^\pi(x,a)(y), \qquad q(y) := (T^\pi\eta)(x,a)(y).
\]
We do not observe exact masses on a fixed grid; instead we:
\begin{enumerate}
    \item Sample from the continuous densities:
    \[
    y^{(1)}, \dots, y^{(N)} \sim p, \qquad \tilde y^{(1)}, \dots, \tilde y^{(M)} \sim q.
    \]
    \item Estimate $p$ and $q$ via KDE on each support using a kernel $K_h$ with bandwidth $h > 0$:
    
    on the predicted support $\{y_i\}_{i=1}^N$,
    \begin{align*}
        \hat p(y_i) &= \frac{1}{N} \sum_{k=1}^N K_h(y_i - y^{(k)}), \\
        \hat q(y_i) &= \frac{1}{M} \sum_{j=1}^M K_h(y_i - \tilde y^{(j)}),
    \end{align*}
    
    on the target support $\{\tilde y_j\}_{j=1}^M$,
    \begin{align*}
        \hat p(\tilde y_j) &= \frac{1}{N} \sum_{k=1}^N K_h(\tilde y_j - y^{(k)}), \\
        \hat q(\tilde y_j) &= \frac{1}{M} \sum_{j=1}^M K_h(\tilde y_j - \tilde y^{(j)}).
    \end{align*}
    
    \item Discretize these KDEs into mass vectors on each grid (e.g.\ by Riemann approximation):
    \[
    w_i^{(y)} \approx \frac{\hat p(y_i)\,\Delta y}{\sum_{k} \hat p(y_k)\,\Delta y},
    \quad
    v_i^{(y)} \approx \frac{\hat q(y_i)\,\Delta y}{\sum_{k} \hat q(y_k)\,\Delta y},
    \]
    and similarly $w^{(\tilde y)}, v^{(\tilde y)}$ on $\{\tilde y_j\}$.
\end{enumerate}
The practical loss we use (Eq. (11) in the main text) is then
\begin{equation}
    \mathcal{L}(\eta^\pi(x,a), T^\pi\eta(x,a))
    \;=\;
    D\big(w^{(y)}, v^{(y)}\big) + D\big(w^{(\tilde y)}, v^{(\tilde y)}\big),
    \label{eq:practical_loss_app}
\end{equation}
where $D$ is exactly the metric defined in \eqref{eq:geom_D_app}, applied separately on the predicted and target supports.
Thus, in practice, we always apply $D$ to KDE-smoothed mass vectors, not to raw one-hot Diracs on the grid.

\subsection{How KDE restores a transport-like behaviour in the one-hot example}

We now revisit the ``two spikes'' example under the KDE-based construction.
Consider two sharply peaked continuous distributions with means $\mu_p$ and $\mu_q$, and treat them as approximations to Dirac masses:
\[
p(y) \approx \delta_{\mu_p}, \qquad q(y) \approx \delta_{\mu_q}.
\]
After applying KDE with bandwidth $h$, we obtain smooth approximations
\[
\hat p(y) \approx K_h(y - \mu_p), \qquad \hat q(y) \approx K_h(y - \mu_q),
\]
and hence, on a grid $\{y_i\}$,
\[
w_i^{(y)} \;\propto\; K_h(y_i - \mu_p), \qquad v_i^{(y)} \;\propto\; K_h(y_i - \mu_q).
\]
Define the difference vector
\[
\Delta_i := w_i^{(y)} - v_i^{(y)}.
\]
The contribution of the predicted support to the loss is
\begin{equation}
    D^2\big(w^{(y)}, v^{(y)}\big) = \frac{1}{N^2} \sum_{i=1}^N \Omega_i \, \Delta_i^2.
    \label{eq:D_KDE_app}
\end{equation}
Now compare two regimes:
\begin{itemize}
    \item \textbf{Near modes:} $|\mu_p - \mu_q| \ll h$.
    The two KDEs $K_h(\cdot - \mu_p)$ and $K_h(\cdot - \mu_q)$ overlap heavily. On most grid points, especially in regions where $\Omega_i$ is large, we have
    \[
    |\Delta_i| = \big|w_i^{(y)} - v_i^{(y)}\big| \approx 0,
    \]
    so the weighted sum $\sum_i \Omega_i \Delta_i^2$ in \eqref{eq:D_KDE_app} is relatively small.

    \item \textbf{Far modes:} $|\mu_p - \mu_q| \gg h$.

The two KDEs overlap very little. There are two separated regions where one of $w_i^{(y)}, v_i^{(y)}$ is large and the other is close to zero, so
\[
|\Delta_i| \approx \max\big\{w_i^{(y)}, v_i^{(y)}\big\}
\]
on those regions. Since these regions typically lie in parts of the grid where $\Omega_i$ is sizeable, many terms $\Omega_i \Delta_i^2$ contribute, making $\sum_i \Omega_i \Delta_i^2$ (and thus $D$) significantly larger than in the near-mode case.
\end{itemize}

As a result, for moderate bandwidth $h$ we have qualitatively
\[
D_{\text{KDE}}(\delta_{\mu_p}, \delta_{\mu_q}) \quad \text{small if } |\mu_p - \mu_q| \ll h,
\qquad
D_{\text{KDE}}(\delta_{\mu_p}, \delta_{\mu_q}) \quad \text{large if } |\mu_p - \mu_q| \gg h.
\]
The same reasoning applies to the symmetric term $D(w^{(\tilde y)}, v^{(\tilde y)})$ on the target support in \eqref{eq:practical_loss_app}. Taken together, this shows that:

While the exact discrete metric $D$ on one-hot mass vectors does not by itself strongly encode ``near vs far'' behaviour, the KDE-based practical loss $\mathcal{L}$ applies $D$ to smoothed mass vectors whose differences $(w_i - v_i)^2$ reflect the spatial separation of modes. As a result, $\mathcal{L}$ behaves more like a transport-style distance: it penalizes discrepancies between distributions increasingly as their mass moves further apart.

In the limit $h \to 0$, the KDEs collapse to Dirac masses on the grid and the behaviour reverts to the discrete one-hot case \eqref{eq:geom_D_onehot}. For the moderate bandwidths used in practice, however, KDE smooths each mode and allows the geometry-aware weights $\Omega_i$ in \eqref{eq:D_KDE_app} to amplify spatially structured discrepancies between the predicted and target return distributions.

\section{C51 tends to learn broader distributions when rewards fall between fixed atoms}

\begin{figure}
    \centering
    \includegraphics[width=0.7\linewidth]{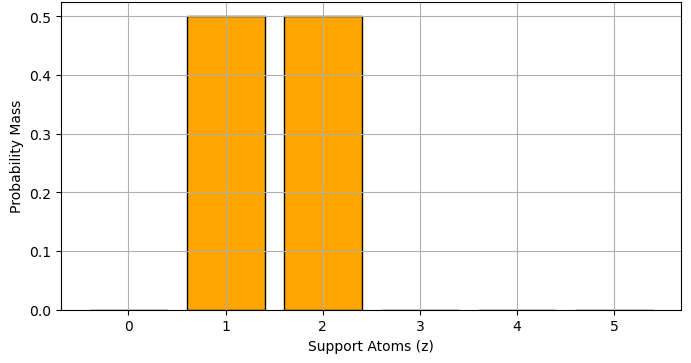}
    \caption{Linear interpolation splits predicted returns among bins in C51}
    \label{fig:projection}
\end{figure}

Consider a situation where the distribution is represented with bins where each bin spans an interval of length 1. Said otherwise, the categorical representation allocates bin 1 to returns between 0 and 1, bin 2 to returns between 1 and 2. Now consider that the return to predict is exactly 1. This value falls right between bin 0 and bin 1. C51 will allocate the same mass to both bins, effectively displaying a wider distribution than it should. This is illustrated in figure \ref{fig:projection} where we used the same mechanism as in C51. We consider $V\in[0,5]$, which gives 6 atoms with $dz=1$. When tring to predict a return of $1.5$, linear interpolation makes it split into 2 bins. The code used to produce this figure is available in the supplementary material. 

\section{Quantile based methods lack control over local spread}

QR-DQN predicts a fixed set of return quantiles (e.g., 5th, 15th, …, 95th percentiles), while IQN learns a function that estimates quantiles for arbitrary probabilities sampled from a uniform distribution. Although IQN provides greater flexibility in querying quantiles, both methods ultimately approximate the return distribution as a discrete set of quantile points. These predicted quantiles are best interpreted as Dirac masses of equal weight, forming a piecewise approximation that omits information about the distribution’s behavior between points. Consequently, quantile-based methods cannot distinguish between smooth distributions and those with sharp peaks or flat regions, nor can they explicitly model local density or control how probability mass is allocated within intervals.

\begin{figure}[htbp]
    \centering
    
    \begin{subfigure}{0.9\columnwidth}
        \centering
        \includegraphics[width=\linewidth]{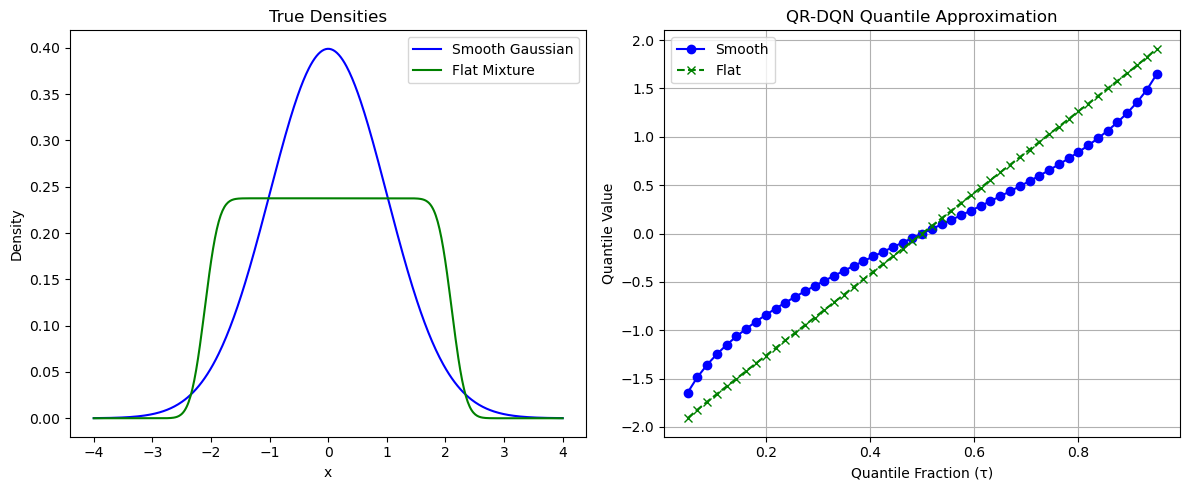}
        \caption{\textbf{Left: }A standard Gaussian and a flat mixture density functions. \textbf{Right: } Their respective quantile approximations using QR-DQN.}
        \label{fig:one_a}
    \end{subfigure}
    
    \vspace{0.5\baselineskip} 
    
    \begin{subfigure}{0.9\columnwidth}
        \centering
        \includegraphics[width=\linewidth]{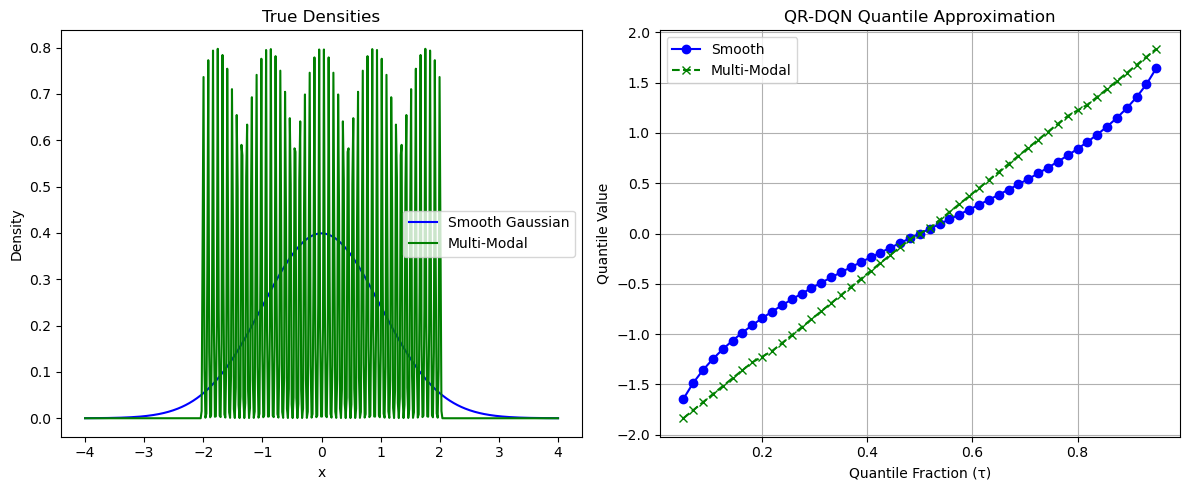}
        \caption{\textbf{Left: } A standard Gaussian and an extremely multimodal distribution. \textbf{Right: } Their respective quantile approximations using QR-DQN.}
        \label{fig:one_b}
    \end{subfigure}
    
    \vspace{0.5\baselineskip}
    
    \begin{subfigure}{0.9\columnwidth}
        \centering
        \includegraphics[width=\linewidth]{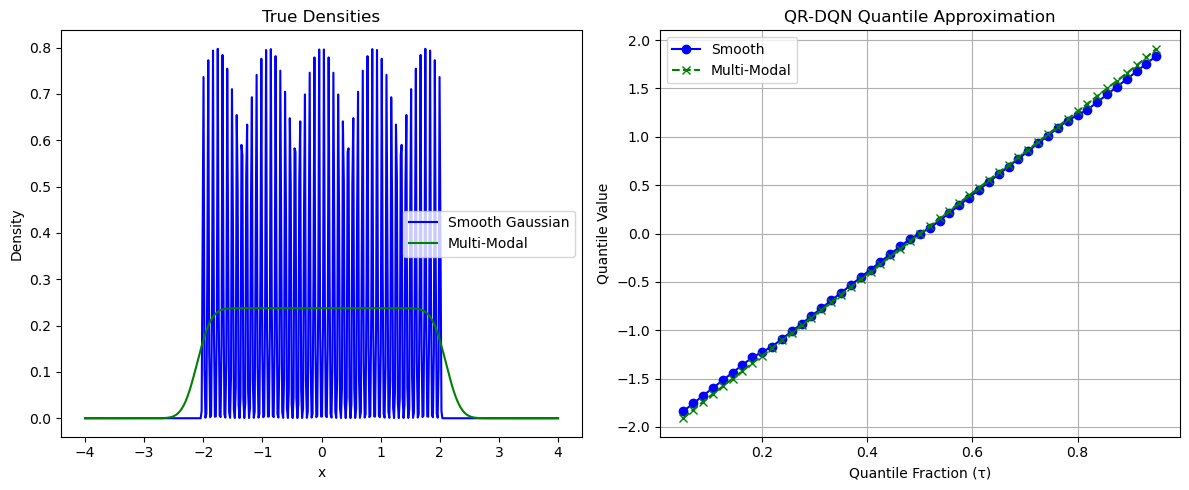}
        \caption{\textbf{Left: } A flat mixture and an extremely multimodal distribution. \textbf{Right: } Their respective quantile approximations using QR-DQN. We observe that they are almost indistinguishable.}
        \label{fig:one_c}
    \end{subfigure}
    
    \caption{QR-DQN quantile approximation for various density functions}
    \label{fig:QR_DQN}
\end{figure}

Our flow-based approach addresses these limitations by modeling a continuous PDF, enabling explicit control over probability mass across the return space and allowing the capture of fine-grained structure and local variability that quantile-based methods cannot. Figure~\ref{fig:QR_DQN} illustrates this difference: three comparisons are shown between a standard Gaussian, a flattened Gaussian, and a peaky distribution. While these distributions are clearly distinct, their quantile approximations are almost identical—in particular, the flattened and peaky cases are nearly indistinguishable. Although this is an extreme example, it effectively demonstrates that quantile-based methods lack the ability to represent local spread.

\section{Additional results}\label{sec:additional}

\subsection{Frozen Lake} 
\begin{figure}[h!]
    \centering
    \includegraphics[width=0.5\linewidth]{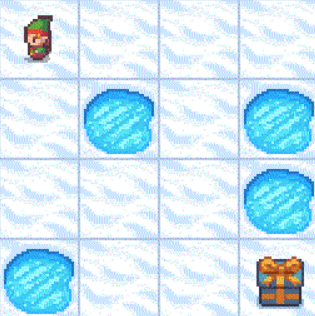}
    \caption{Frozen lake environment}
    \label{fig:frozen_env}
\end{figure}

\begin{figure}[h!]
    \centering
    \includegraphics[width=0.6\textwidth]{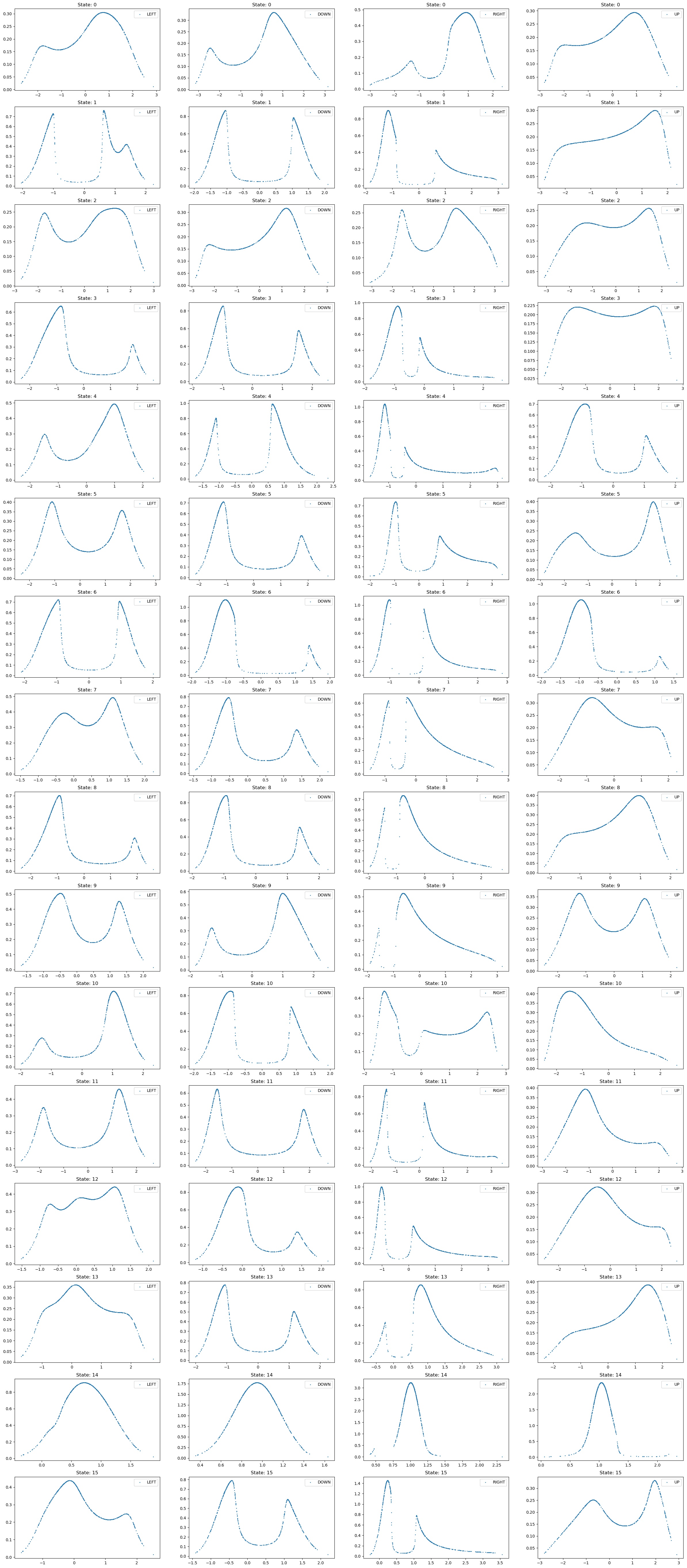}
    \caption{Learnt value distributions using the surrogate loss for each state-action pair of the Fozen Lake environment. States are numbered from left to right and up to down, i.e upper left is 0, bottom right is 16.}
    \label{fig:frozen_lake}
\end{figure}

Figure~\ref{fig:frozen_env} illustrates the Frozen Lake environment, a stochastic grid-world designed to highlight multimodal value distributions. An agent must navigate from a starting point to a goal while avoiding holes in the ice. Due to slipperiness, intended moves may result in perpendicular actions; for example, choosing to move right results in a 1/3 chance of moving right, up, or down. This high level of stochasticity induces significant variability in returns, making Frozen Lake a suitable testbed for visualizing multimodal Q-value distributions. The learned value distributions for each state-action pair, obtained using the surrogate loss, are displayed in Figure~\ref{fig:frozen_lake}. States are ordered left to right, top to bottom.

\subsection{ATARI-5 Raw scores}
\begin{table}[h!]
\centering
\caption{Raw scores on ATARI-5 Benchmark.}
\resizebox{\textwidth}{!}{%
\begin{tabular}{lcccccccc}
\toprule
\textbf{Games} & \textbf{Random} & \textbf{Human} & \textbf{DQN} & \textbf{C51} & \textbf{IQN} & \textbf{NFDRL-E} & \textbf{NFDRL-S}\\
\midrule
Battle Zone   & 2,360.0 & 37,187.5 & 29,900.0 & 28742.0   & \textcolor{blue}{\textbf{42244.0}} & 32800.0 & \textbf{34540.0} \\
Double Dunk  & -18.6   & -16.4 & -6.6  & 2.5  & 5.6 & \textcolor{blue}{\textbf{7.8}} & 7.6 \\
Name this Game & 2,292.3 & 8,049.0  & 8,207.8 & 12,542.0 & \textcolor{blue}{\textbf{22,682.0}} & 15,667.2 & \textbf{17,309.5}\\
Phoenix & 761.4 & 7,242.6 & 8,485.2 & 17,490 & \textcolor{blue}{\textbf{56,599}} & 18,914 & \textbf{20,042}\\
Q*Bert & 163.9 & 13,455.0 & 13,117.3 & 23,784 & 25,750 & 22,671 & \textcolor{blue}{\textbf{25,852}}\\
\bottomrule
\end{tabular}
}
\label{tab:raw_scores}
\end{table}

\subsection{Implemetation Details and hyperparameters}\label{sec:hyperparams}

\paragraph{Architecture.} We base all baselines and our method on the same underlying neural network. Its architecture and training loop composition follows the structure used in CleanRL \citep{cleanRL}. Our method however requires 4 separate heads outputing $\{w_i^j,\mu_i^j,\sigma_i^j\}$ and $G^j_{\text{max}}$. Hyperparameters are displayed in table \ref{tab:hyperparameters}. Our model was trained on an Nvidia RTX A5000 GPU.

As mentioned in main paper section 3.4, we choose $\sigma = 0.05$ to ensure the resulting Gaussian is sharply peaked around $r$, while still numerically stable. This value is derived by matching the resolution of the C51 model, which discretizes the support $[0, 10]$ into 51 bins, yielding a bin width of approximately $0.2$. To ensure that 95\% of the Gaussian’s mass lies within the central bin (i.e., $[r - 0.1, r + 0.1]$), we solve $2\sigma = 0.1$, yielding $\sigma = 0.05$. Of course, different values can be chosen to allow for sharper distributions.

\paragraph{Action selection.} During training, actions are chosen using an $\epsilon$-greedy strategy: with probability $\epsilon$, a random action is selected; otherwise, the model selects the best action. The value of $\epsilon$ decreases over time. In the latter case, action selection follows the approach of C51 or IQN. The model includes one head per possible action, each outputting the parameters of a Gaussian mixture and $G^{max}$—defining a flow function per action. From these, $n$ samples are drawn from the base distribution and passed through the corresponding flow. This yields $n$ return values and densities per action. The expected return is computed for each, and the action with the highest expected value is chosen.

\begin{table}[h!]
\centering
\begin{tabular}{ll}
\toprule
\textbf{Hyperparameter} & \textbf{Value / Description} \\
\midrule
\texttt{total\_timesteps} & 10,000,000 \quad (total timesteps of the experiments) \\
\texttt{learning\_rate} & 5e-5 \quad (learning rate of the optimizer) \\
\texttt{max\_norm} & 3.0 \quad (maximum allowable gradient norm) \\
\texttt{num\_envs} & 4 \quad (number of parallel game environments) \\
\texttt{buffer\_size} & 1,000,000 \quad (size of the replay memory buffer) \\
\texttt{gamma} & 0.99 \quad (discount factor) \\
\texttt{target\_network\_frequency} & 1 \quad (steps between target network updates) \\
\texttt{batch\_size} & 64 \quad (batch size from replay memory) \\
\texttt{start\_e} & 1.0 \quad (starting epsilon for exploration) \\
\texttt{end\_e} & 0.01 \quad (final epsilon for exploration) \\
\texttt{exploration\_fraction} & 0.2 \quad (fraction of timesteps for epsilon decay) \\
\texttt{learning\_starts} & 30,000 \quad (timestep to start learning) \\
\texttt{train\_frequency} & 4 \quad (training frequency) \\
\texttt{hidden\_size\_1} & 512 \quad (size of first hidden layer) \\
\texttt{hidden\_size\_2} & 256 \quad (size of second hidden layer) \\
\texttt{n\_flows} & 1 \quad (number of flows) \\
\texttt{n\_components} & 4 \quad (number of components in the mixture) \\
\texttt{n\_samples} & 500 \quad (samples drawn from the base distribution) \\
\texttt{final\_reward\_variance} & 0.1 \quad (final state reward normal distribution variance)\\
\texttt{bandwidth} & 0.05 \quad (KDE bandwidth)\\
\bottomrule
\end{tabular}
\caption{List of hyperparameters used in our experiments.}
\label{tab:hyperparameters}
\end{table}



\end{document}